\documentclass{article}

     \PassOptionsToPackage{numbers, compress}{natbib}

     \usepackage[final]{neurips_2022}

\usepackage[utf8]{inputenc} 
\usepackage[T1]{fontenc}    
\usepackage[hidelinks]{hyperref}
\usepackage{url}            
\usepackage{booktabs}       
\usepackage{amsfonts}       
\usepackage{nicefrac}       
\usepackage{microtype}      
\usepackage{xcolor}         

\usepackage{amsmath}
\usepackage{amssymb}
\usepackage{mathtools}
\usepackage{amsthm}

\usepackage{algorithm, algorithmic}

\usepackage{subcaption}

\usepackage{wrapfig}

\usepackage[capitalize,noabbrev]{cleveref}

\usepackage{multirow}

\usepackage{wrapfig,lipsum,booktabs}

\renewcommand\vec{\boldsymbol}

\newcommand\bx{\vec{x}}
\newcommand\by{\vec{y}}

\newcommand\bv{\vec{v}}
\newcommand\bb{\vec{b}}
\newcommand\bB{\vec{B}}
\newcommand\bA{\vec{A}}

\newcommand\bxi{\bx_i}
\newcommand\byi{\by_i}
\newcommand\bxj{\bx_j}
\newcommand\byj{\by_j}

\newcommand\btheta{\vec{\theta}}

\newcommand\bphi{\vec{\phi}}

\title{Remember the Past: Distilling Datasets \\ into Addressable Memories for Neural Networks}

\author{%
  Zhiwei Deng\quad 
    Olga Russakovsky \\
    Department of Computer Science\\
    Princeton University \\
    \{zhiweid, olgarus\}@cs.princeton.edu
}

\begin{document}

\maketitle

\begin{abstract}
We propose an algorithm that compresses the critical information of a large dataset into compact addressable memories. These memories can then be recalled to quickly re-train a neural network and recover the performance (instead of storing and re-training on the full original dataset). Building upon the dataset distillation framework, we make a key observation that a \textit{shared common representation} allows for more efficient and effective distillation. Concretely, we learn a set of bases (aka ``memories'') which are shared between classes and combined through learned flexible addressing functions to generate a diverse set of training examples. This leads to several benefits: 1) the size of compressed data does not necessarily grow linearly with the number of classes; 2) an overall higher compression rate with more effective distillation is achieved; and 3) more generalized queries are allowed beyond recalling the original classes. 
We demonstrate state-of-the-art results on the dataset distillation task across six benchmarks, including up to 16.5\% and 9.7\% in retained accuracy improvement when distilling CIFAR10 and CIFAR100 respectively. We then leverage our framework to perform continual learning, achieving state-of-the-art results on four benchmarks, with 23.2\% accuracy improvement  on MANY. 
The code is released on our project webpage\footnote{\url{https://github.com/princetonvisualai/RememberThePast-DatasetDistillation}}. 




\end{abstract}




\section{Introduction}

Compressing a large amount of information into a small memory storage space is one of the key components of human intelligence~\cite{brady2009compression, loftus2019human, anderson2014human} -- a person can retrieve memories from the past and quickly recover the corresponding skills. Deep learning methods have made large strides in building task-specific models, but are shown to easily forget past knowledge when learning new tasks~\cite{goodfellow2013empirical,mccloskey1989catastrophic}. 


To equip neural network learners with memorizing ability, dataset distillation~\cite{wang2018dataset} is proposed as a potential solution. Concretely, a compressed set of examples (memories) is learned to summarize the key information in a dataset that affects model training; these examples can then be used to quickly retrain models and recover the corresponding skills. This differs from the standard reconstruction-based compression algorithms~\cite{goodfellow2014generative, gemp2020eigengame, kingma2013auto} and shows strong performance~\cite{zhao2021DC,zhao2021DSA,nguyen2021dataset,nguyen2021iclr}.


A critical question in building powerful compressed memories is: what structures and representations should we use to build the memories? An effective structure and organization of memories can lead to different fundamental assumptions about data and affect the compression and learning behaviours. Existing works~\cite{zhao2021DC,zhao2021DSA,zhao2021dataset,nguyen2021dataset,cazenavette2022dataset,nguyen2021iclr,wang2018dataset} follow a simple representation, where a set of learnable examples is assigned for each class. However, under this assumption, the size of the memories can linearly grow with the number of classes, making the distillation of datasets with a large number of classes challenging. Naturally, this can potentially lead to redundancies in the learned memories, due to the separation of data among classes. Furthermore, this representation is less generalizable to continuous label space, where infinite number of label values exists.

In our paper, we make the observation that there is information shared between classes, and hypothesize that a common and compact representation exists for all classes. Following this hypothesis, we propose to formulate the problem as a memory addressing process, where the memories store a common set of bases shared by all classes, and the recombination of bases is performed through an addressing function. This decomposition between memories and addressing functions enables the possibility that all common information is stored in one part of the representation, and the accessing of the common information depends on the specific labels and is handled through an extra function. We find that this formulation can significantly improve both the compression rate and the performance. 


We adopt the back-propagation through time learning framework to train the memories and addressing functions, and identify several critical factors that can improve the performance. Specifically, we find that adopting the momentum term, and performing long unrolls in the inner optimization loop are both critical. This differs from the common usage of bi-level optimization algorithm on this task~\cite{wang2018dataset, sucholutsky2019soft}, and leads to strong performance outperforming single-step gradient matching methods~\cite{zhao2021DC,zhao2021DSA} even with the simple data representation.

In the experiments, we extensively evaluate our algorithm on six benchmarks of the Dataset Distillation task, and show that it consistently outperforms previous state-of-the-art by a significant margin. For example, we achieve 66.4\% accuracy on CIFAR10 with the storage space of $1$ image per class, improving over the previous state-of-the-art KIP method~\cite{nguyen2021dataset,nguyen2021iclr} by 16.5\%. We further demonstrate our method on the continual learning tasks, and show that a simple ``compress-then-recall'' method using our framework leads to state-of-the-art results on four datasets. For example, we outperform all prior methods by 23.2\% in retained accuracy on the challenging MANY~\cite{riemer2018learning} benchmark. Finally, we demonstrate the generality of our approach by extending to image-based (rather than label-based) memory recall, and synthesizing new classifiers (unseen during training) from our distilled memories. 


\section{Related works}

\textbf{Dataset Distillation.} The task of dataset distillation is fundamentally a compression problem, with a different prioritization on the information contained in data. There have been several lines of methods, developed with different criteria to prioritize information. \textit{Generalization loss} with bi-level optimization framework~\cite{finn2017model,franceschi2017forward,raghu2020teaching} has been widely studied and is used in the early works of dataset distillation~\cite{wang2018dataset,sucholutsky2019soft}. It emphasizes on the loss at the final optimization state. \textit{Gradient-matching or score-matching} methods~\cite{zhao2021DC,zhao2021DSA,cazenavette2022dataset} are adopted to directly match the induced gradients from synthetic data. If ideally matched over the gradient field, the compressed dataset can naturally lead to the same model parameters with gradient descent. \textit{Kernel method}~\cite{nguyen2021iclr,nguyen2021dataset} shows that with the connection to Gaussian processes, a kernel inducing points method can be used to achieve strong performance, but with large computation costs. These are also connected with the recent progress on pragmatic compression methods~\cite{reddy2021pragmatic,zhao2021comparing}, which compress or match distributions based on a decision process (in dataset distillation's case, the gradient descent search process).


\textbf{Continual learning.} Broadening the learning paradigms, continual learning problem aims to build agents that learn through a stream of tasks and accrue knowledge along the process. ``Catastrophic forgetting''~\cite{mccloskey1989catastrophic, ratcliff1990connectionist,goodfellow2013empirical} is a well-known phenomenon in this setting, where the neural network forgets previous skills when learning new ones. Various methods~\cite{lopez2017gradient, chaudhry2019tiny,veniat2020efficient,nguyen2017variational,riemer2018learning,gupta2020maml,rebuffi2017icarl,von2021learning,prabhu2020gdumb,de2019continual,hadsell2020embracing,rusu2016progressive,yoon2018lifelong} on regularization, replay, or dynamic model, have been proposed to alleviate the issue and address the ``stability-plasticity dilemma''~\cite{mermillod2013stability,mirzadeh2020understanding}. Memory buffer has been a critical component in the past methods~\cite{lopez2017gradient, chaudhry2019tiny,veniat2020efficient,nguyen2017variational,riemer2018learning,gupta2020maml,saha2020gradient,derakhshani2021kernel}, but mainly relies on a random selection of real samples with different strategies. Recently, several works extend the usage of memory to storing random basis~\cite{derakhshani2021kernel} (online setting) or SVD bases~\cite{saha2020gradient} (offline setting).


\section{Background: dataset distillation}


The task of Dataset Distillation~\cite{wang2018dataset} is proposed to compress the key information of a large-scale training dataset into a small amount of learned data, which can be stored using limited memory space and retrieved through label indices or task information to recover the performance of a model.

\textbf{Problem Setting.} Formally, given a large dataset $\mathcal{D}_{tr}=\{(\bxi, \byi)\}_{i=1}^{N}$ containing $N$ pairs of training data $(\bxi, \byi)$, where $\bxi$ is an image and $\byi$ is the corresponding label in $C$ classes, a small dataset $\mathcal{D}_{s}=\{(\bxj^{\prime}, \byj)\}_{j=1}^{N^{\prime}}, N^{\prime} \ll N,$ can be synthesized or distilled, such that a model trained on $\mathcal{D}_s$ can have the same generalization ability as ones trained on $\mathcal{D}_{tr}$:
\begin{equation}
\label{eqn:obj}
    \mathbb{E}_{(\bx, \by)\sim \mathcal{D}_{te}}\big[\mathbf{m}(f(\bx;\btheta^{(*)}), \by)\big] \simeq \mathbb{E}_{(\bx, \by)\sim \mathcal{D}_{te}}\big[\mathbf{m}(f(\bx;\btheta^{\prime(*)}), \by)\big]
\end{equation}
where $\btheta^{(*)}$ and $\btheta^{\prime(*)}$ are the optimized parameters using $\mathcal{D}_{tr}$ and $\mathcal{D}_{s}$ respectively, $\mathbf{m}$ is a metric, e.g., accuracy, and $\mathcal{D}_{te}$ is the test dataset. The model is often a neural network classifier $f(\cdot;\btheta)$ parameterized by $\btheta$ and trained with a loss function $\ell(f(\bx;\btheta), \by)$. 

\textbf{Synthetic dataset representations.} The synthetic dataset contains the core information that needs to be learned. The representation of the synthetic data affects the compactness and effectiveness of the distillation process. In existing methods~\cite{wang2018dataset,zhao2021DC,zhao2021DSA,zhao2021dataset,nguyen2021dataset,cazenavette2022dataset}, the dataset $\mathcal{D}_s$ is defined as a collection of learnable data samples $(\bx^{\prime}, \by)$, and the number of samples is separately and equally distributed across classes. This representation has several disadvantages: first, the number of synthetic data samples needed for a dataset grows linearly with the number of classes, leading to limited applicability when the number of classes is large or undefined (e.g., language or other continuous labels); second, the potentially shared and common information across classes is ignored -- this results in a less compact representation of the distilled information and lower compression rate; lastly, the representation is not able to generalize to new classes or tasks, due to the lack of common representation learned across classes.



\section{Model}

\textbf{Overview.} In this section, first, we present a new perspective of the problem, where the Dataset Distillation problem is formulated as a \textit{memory addressing process}: instead of learning synthetic images separately for each class, we construct and learn a common memory representation that can be accessed through addressing matrices to construct synthetic datasets. Under this formulation, the number of synthetic images does not need to grow linearly with the number of classes, the shared information among classes can be exploited to reduce redundancies and improve compression rate, and datasets can be distilled with respect to more generic queries. Second, we further show several critical empirical facets of back-propagation through time framework, which lead to drastic improvements on the performance and outperform the single-step gradient matching methods. This is in contrast with the current common observation that gradient matching outperforms back-propagation through time framework on dataset distillation tasks.

In the following, we present the two core components of our method, (1) the new formulation of dataset distillation, with memories and addressing matrices in Sec.~\ref{sec:mem}, and (2) the learning framework under back-propagation through time in Sec.~\ref{sec:learn-alg}.

\begin{figure}[t]
  \centering
  \includegraphics[width=1.0\linewidth]{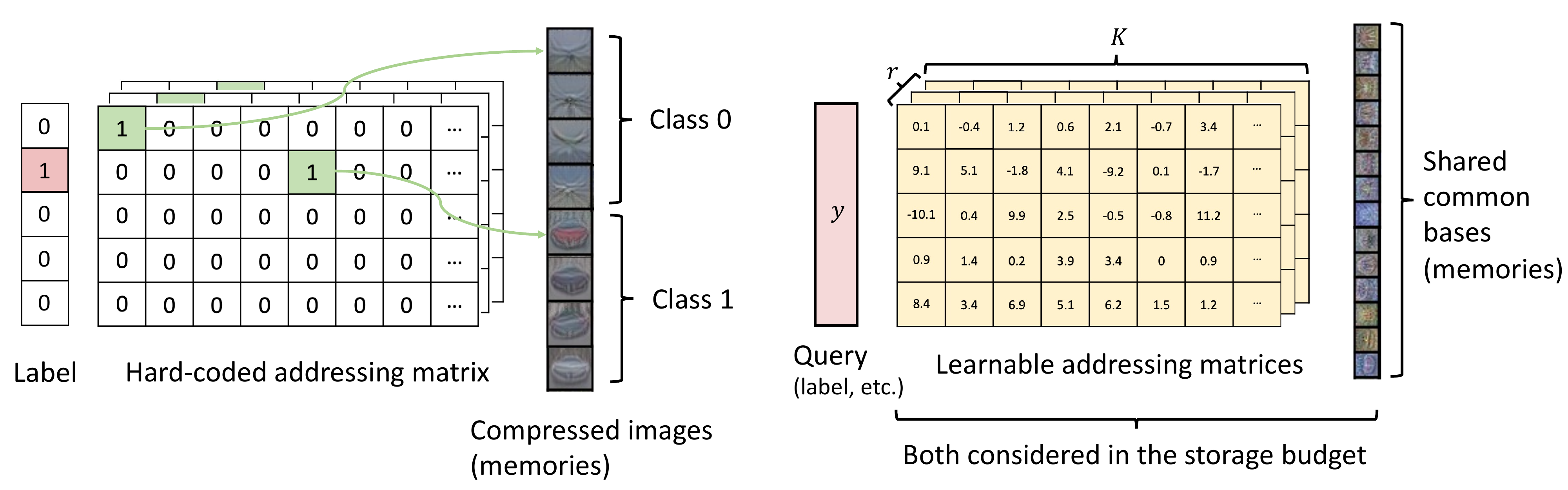}
   \caption{Distilling a large-scale dataset into compressed memories. \textit{Left:} the standard dataset distillation task under the formulation of memory addressing. The addressing matrices are hard-coded with $1$s and $0$s to fetch the corresponding compressed image in memories. The memory size grows linearly with number of classes. \textit{Right:} learnable addressing matrix with shared common bases (number of bases can be flexibly defined). The information sharing between classes is captured in this representation. The queries can be generalized to any vector representation, besides one-hot labels, i.e. for a general dataset from $\mathcal{X}$ to $\mathcal{Y}$, it can be distilled into memories for recall and model re-training.}
   \label{fig:main_model}
   \vspace{-5mm}
\end{figure}

\subsection{Dataset Distillation as memory addressing}
\label{sec:mem}

\textbf{Problem formulation.} Given a task-specific dataset $\mathcal{D}_{tr}=\{\bxi, \byi\}_{i=1}^{N}, \bx \in \mathcal{X}, \by \in \mathcal{Y}$, we aim to learn a single compact and compressed representation $\mathcal{M}$, referred to as memories, that can be accessed through a learned addressing function $\mathcal{A}(\cdot)$. $\mathcal{A}(\cdot)$ takes all possible values of $\by$ as input and recalls the corresponding synthetic data. With a set of $\{\byi\}$, a synthetic dataset recalled using the above process can train a model $f_{\btheta}: \mathcal{X}\rightarrow
 \mathcal{Y}$ from scratch and obtain the same generalization ability as trained on $\mathcal{D}_{tr}$.

Ideally, the memories $\mathcal{M}$ and the addressing function $\mathcal{A}$ can jointly capture the critical information that defines the task mapping from $\mathcal{X}$ to $\mathcal{Y}$, such that the recalled synthetic data given a query $\byi$ contains distinctive information that defines $\byi$ and the synthetic dataset recalled with $\{\byi\}$ can satisfy eqn.~\ref{eqn:obj} when used for re-training. 
For example, in the standard classification tasks, we can enumerate all possible values in the label space (discrete) and address the memories, to construct the synthetic dataset that contains critical information. Under this formulation, since we are learning a single, shared and accessible representation for all $\by$s, the size of memories can be defined flexibly regardless of the number of classes, removing the linear growth limitation in the standard distillation settings. There is also no constraint on the form of $\by$s, which can be either discrete or continuous. The \textit{storage budget} or \textit{compression rate} is calculated by considering the storage space of parameters in both memories and addressing functions, which should be as compact as possible.

\textbf{Memory representation.} We use a set of bases to store in the memories $\mathcal{M}=\{\bb_1,...,\bb_K\}$, where each vector $\bb_k \in \mathbb{R}^d$ has the same dimension as $\bx \in \mathbb{R}^d$, and all vectors collectively define the intrinsic components in a dataset that characterizes the task mapping from $\mathcal{X}$ to $\mathcal{Y}$. Through re-using the bases, we can produce a desired synthetic dataset for model re-training. \textit{Spatial redundancies in images:} note that, as a special case, images can contain redundancies spatially and be stored in a downsampled version to improve the compression rate. The downsampled image bases can be passed via a deterministic upsampling process (e.g., bilinear interpolation) to recover the original resolution.

\textbf{Memory addressing.} For each query $\by$, we use a parameterized function $\mathcal{A}(\by)$ to re-combine the bases in the memories $\mathcal{M}$. Similar to previous methods on accessing memories, we use $\mathcal{A}$ to produce a set of coefficients, and linearly combine the bases to produce synthetic data. Formally, to retrieve $r$ synthetic examples for each $\by$, we define a set of matrices $\{\bA_1, ..., \bA_r\}, \bA_i \in \mathbb{R}^{d_y \times K}$, where $d_y$ is the dimension size of $\by$ and $r$ is the number of data samples that can be retrieved. With the memories $\mathcal{M}=\{\bb_1,...,\bb_K\}$, we define:
\begin{equation}
\label{eqn:addr}
    \bx^{\prime T}_i = \by^T \bA_i [\bb_1;...;\bb_K]^T, \bx^{\prime} \in \mathbb{R}^{d\times 1}
\end{equation}
where $\by \in \mathbb{R}^{d_y \times 1}$ is in a vectorized form, such one-hot encoding of categorical labels, and $\bv=\by^T \bA_i$ corresponds to a coefficient vector $\bv$ that combines the bases. The produced synthetic data $\bx^{\prime}$ is paired with $\by$ as the corresponding label. Our model is shown on the right of figure~\ref{fig:main_model}.

\textbf{Constructing a dataset.} To construct a synthetic dataset $\mathcal{D}_s$ for model re-training, we are often given a set of samples $\{\byi\}$, or can enumerate all possible values of $\by$ (if discrete). With the set $\{\byi\}$, we use eqn.\ref{eqn:addr} to address and retrieve the synthetic dataset $\mathcal{D}_s^{\by=\byi}=\{(\bx^{\prime}_j, \byi)\}_{j=1}^{r}$ for each $\byi$. The final dataset is the union of all $\mathcal{D}_s=\bigcup \mathcal{D}_s^{\by=\byi}$. The dataset $\mathcal{D}_s$ can be used in either a minibatch form for stochastic gradient descent, or as a whole for batch gradient descent.

\textbf{Generalized possibilities of queries.} Another advantage of our formulation, besides a compact and shared representation, is the various possibilities of queries $\by$. In principle, under this formulation, the label $\by$ can be flexibly defined as other forms, such as language or audio, where the representation resides in a continuous space or follows a distribution $p(h(\by))$ defined by a feature extractor $h(\cdot)$. The set of $\{\byi\}$ then can be sampled from the distribution $p$ instead of having to enumerate all possible values. This provides a general way of compressing or distilling a large dataset without constraint on the forms of labels. 

\textbf{Connection with standard Dataset Distillation.} In the standard setting, dataset distillation is defined for classification tasks with discrete labels and each label owns its unique set of synthetic data. We can show that this is a special case of our formulation: if the bases are constructed as the collection of those label-specific synthetic data ($K=N^{\prime}$, $N^{\prime}$ is the total size of the synthetic dataset), the addressing matrices $\bA_i$ are defined in space $\{0,1\}^{C\times N^{\prime}}$ where $\bA_i[m,n]=1$ if $n$ equals to $m(N^{\prime} / C) + i$ ($i^{th}$ item of $m^{th}$ class) and $\bA_i[m,n]=0$ at other positions, then the "retrieval" process can also be defined as eqn.~\ref{eqn:addr}.

\subsection{Learning framework: back-propagation through time}
\label{sec:learn-alg}
In this section, we build upon the back-propagation through time algorithm and discuss in detail the learning framework that performs the distillation process from a dataset to memories and addressing functions. 


Starting from notations, we define the parameters contained in both memories and addressing functions as $\bphi$, which are collectively optimized. A loss function is $\ell(\cdot,\cdot)$ is defined on a task-specific dataset. We denote an optimization algorithm as $\textsc{opt}(\cdot,\cdot;\alpha,\beta, \ell, T)$, where $\alpha$ and $\beta$ are the learning rate and the momentum rate, respectively, and $T$ is the number of optimization steps. For a single step optimization, we denote it as $\textsc{opt-step}(\cdot,\cdot;\alpha,\beta)$.

\begin{algorithm}[tb]
   \caption{}
   \label{alg:addr}
   \begin{algorithmic}[1]
   \STATE {\bfseries hyperparameters:} Momentum rate $\beta_{0}, \beta_{1}$, learning rate $\alpha_{0}, \alpha_{1}$ for $\btheta$ and $\bphi$ respectively.
   \STATE {\bfseries input:} Dataset $\mathcal{D}_{tr}$, memories $\mathcal{M}$, addressing function $\mathcal{A}$, loss function $\mathcal{\ell(\cdot,\cdot)}$
   \REPEAT
   \STATE Sample a subset of labels $\mathcal{Y}^{\prime}$ 
   \STATE Address the memories $\mathcal{M}$ and obtain synthetic dataset $\mathcal{D}_{s}^{\mathcal{Y}^{\prime}}$ with eqn.~\ref{eqn:addr} \\  
   \STATE Randomly initialize model parameters $\btheta_0$ \\
   \STATE Initialize momentum $\vec{m}_0 =0$\\
   \FOR{$t=1$ {\bfseries to} $T$}
   \STATE Sample a minibatch $\bB_{s}=\{(\bxi^{\prime}, \byi)\}$ from $\mathcal{D}_{s}^{\mathcal{Y}^{\prime}}$
   \STATE Compute $\mathcal{L}=\frac{1}{|\bB_s|}\sum_{i=1}^{|\bB_s|} \ell (f_{\btheta_{t-1}}(\bxi^{\prime}), y_{i})$
   \STATE \textbf{Update momentum} $\vec{m}_t =\beta_{0}\vec{m}_{t-1} + \frac{d \mathcal{L}}{d\vec{\theta}_{t-1}}$
   \STATE Update $\vec{\theta}_t=\vec{\theta}_{t-1} - \alpha_{0}\vec{m}_t$
   \ENDFOR
   \STATE Sample a minibatch $B=\{(\bxi, \byi)\}$ from $\mathcal{D}_{tr}$ with labels in $\mathcal{Y}^{\prime}$
   \STATE Compute $J(\vec{\phi})=\frac{1}{|\bB|}\sum_{i=1}^{|\bB|} \ell(f_{\btheta_T}(\bxi), \byi)$ \\
   \STATE Update $\bphi\leftarrow \textsc{opt-step}(\bphi, J(\vec{\phi}), \alpha_{1}, \beta_{1})$ \\
   \UNTIL{Converge}
\end{algorithmic}
\end{algorithm}

To learn the parameters $\bphi=\{\mathcal{M}, \mathcal{A}\}$, we follow a standard bi-level optimization framework with back-propagation through time (BPTT), where the inner-loop uses the synthetic dataset $\mathcal{D}_s$ to train a randomly initialized model starting from scratch, and a generalization loss is computed using a minibatch $\bB=\{(\bxi, \byi)\}$ sampled from $\mathcal{D}_{tr}$. The parameters $\bphi$ are implicitly contained in the synthetic dataset $\mathcal{D}_s$ and optimized when minimizing the generalization loss. The bi-level optimization defines: 
\begin{eqnarray}
\label{eqn:learning}
    \min J(\bphi) &=& \frac{1}{|\bB|}\sum_{i=1}^{|\bB|} \ell(f(\bxi;\btheta^*), \byi), \nonumber \\
    \text{subject to \quad} \btheta^* &=& \textsc{opt} \big(\btheta_0, \mathcal{D}_{s};\alpha_{0}, \beta_{0}, \ell, T\big)
\end{eqnarray}
where $\btheta_0$ represents the initializing parameters, $\btheta^{*}$ is the optimized model parameters in the inner optimization loop, $\alpha_{0}$ and $\beta_{0}$ are the learning rate and momentum rate for $\text{opt}(\cdot,\cdot)$, and $J(\bphi)$ is the generalization loss on minibatch $\bB$. In practice, for each inner loop training, we randomly sample a subset $\mathcal{Y}^{\prime}$ from $\by$s and retrieve the corresponding subset $\mathcal{D}_{s}^{\mathcal{Y}^{\prime}}$. This reduces the computation cost in inner loops. We empirically observe that equivalent results can be achieved with faster runtime. The algorithm is summarized in Alg.~\ref{alg:addr}, where lines 8-13 define the inner loop optimization process $\textsc{opt}(\cdot,\cdot;\alpha,\beta,\ell,T)$, and the gradients of generalization loss (line 15) is back-propagated through the inner loop to update $\bphi$. Note that, in principle, the inner loop optimization can be performed using any optimizer. In this paper, we mainly rely on the standard stochastic gradient descent with momentum to train the distilled data. 


\textbf{Critical factors in BPTT.} Although being a natural choice in performing dataset distillation and adopted in the original work~\cite{wang2018dataset}, the BPTT framework has been shown to underperform other algorithms, such as single-step gradient matching methods~\cite{zhao2021DC, zhao2021DSA}, on various benchmarks. The underlying causes that hinder the performance of the algorithm are still underexplored. In our work, we investigate and identify the factors that can unleash the potential of back-propagation through time framework on dataset distillation and lead to significant performance boosts.


\textit{Momentum term.} In previous dataset distillation works~\cite{wang2018dataset,sucholutsky2019soft}, the usage of back-propagation through time framework omits the momentum term in the inner loop optimization. Indeed, this has been a common practice in meta-learning tasks~\cite{finn2017model,nichol2018reptile}. Adding momentum terms in meta-learning can potentially even hurt the performance and lead to less gradient diversity~\cite{nichol2018reptile}. However, we observe that, in dataset distillation tasks, the momentum term is crucial for making BPTT excel, even in the relatively short inner loop optimization settings (e.g. 10 or 20 steps). We provide results and analysis in the experiments section.


\textit{Long unrolled trajectories.} Another aspect in using BPTT in dataset distillation is the length of unrolled optimization trajectories in the inner loops. The previous usage of BPTT on this task~\cite{wang2018dataset,sucholutsky2019soft} adopts relatively short inner loop optimization trajectories (e.g. 10-30 steps). Instead, we show that unrolling the trajectories long enough (e.g. 200 steps) with momentum terms can potentially produce $\btheta^{*}$ that better summarizes the information contained in memories and addressing matrices, generating more effective gradients to learn the compressed representation.




\section{Experiments}
We thoroughly evaluate our model and demonstrate the benefits over previous methods. In section~\ref{sec:DD}, we show that using a shared representation is critical to improving the distillation performance and compression rates. Specifically, we observe that there is strong evidence that there is information re-using across classes. We further show the benefits of our model on standard continual learning tasks in section~\ref{sec:cl}. For example, we observe that a simple ``compress-then-recall'' method can achieve performance outperforming state-of-the-art continual learning models with complex designs. Finally, in section~\ref{sec:syn_cls}, we show that storing the compressed data enables synthesizing new classifiers (section~\ref{sec:new_task}) and the shared representation formulation allows more general queries (section~\ref{sec:img_addr}), which can be continuous (e.g. image features).

\subsection{Dataset Distillation}
\label{sec:DD}
In this section, we follow the standard setting of dataset distillation, and perform dataset compression that can be recalled with discrete class labels.

\textbf{Datasets.} We test our models on six standard dataset distillation benchmarks: MNIST~\cite{deng2012mnist}, FashionMNIST~\cite{xiao2017fashion}, SVHN~\cite{netzer2011reading}, CIFAR10~\cite{krizhevsky2009learning}, CIFAR100~\cite{krizhevsky2009learning}, and TinyImageNet~\cite{le2015tiny}. MNIST contains 10 classes with 60,000 writing digit images as training and 10,000 as test set. The images are gray-scale with a shape of $28\times 28$. FashionMNIST is a dataset with clothing and shoe images and consists of a training with size 60,000 and a test set with size 10,000. Each image is $28\times 28$ in gray scale, and has a label from 10 classes. SVHN contains street digit images where each image has shape $32\times 32 \times 3$. The dataset contains 73,257 images for training and 26,032 images for testing. CIFAR10 and CIFAR100 are color image datasets, with 50,000 training images and 10,000 testing images on each. CIFAR10 has 10 classes with 5,000 images per class, and CIFAR100 has 100 classes with 500 images per class. TinyImageNet~\cite{le2015tiny} contains 200 categories with images of resolution 64x64. The training and testing sets have 100,000 and 10,000 images respectively.

\textbf{Experiment settings.} We evaluate our distillation models under three different memory budgets for each dataset: 1/10/50 images per class. We focus on high compression rate scenarios and consider the 1 and 10 settings for CIFAR100. Following previous works~\cite{zhao2021dataset, zhao2021DSA, nguyen2021dataset, cazenavette2022dataset}, the main network architecture used in experiments is a simple convolutional network (ConvNet) with $3\times 3$ filters, InstanceNorm, ReLU and $2\times 2$ average pooling. For the evaluation protocol, each model is evaluated on 20 randomly initialized models, trained for 300 epochs on a synthetic dataset, and tested on a held-out testing dataset. We use one GPU per experiment run. 

\textbf{Memory budget calculation.} Since our model uses memories and addressing matrices to store the compressed information, we treat the total number of images as a memory storage budget. When comparing with $N$ images for $C$ classes, we ensure
\begin{equation}
\label{eqn:budget-eqn}
    \text{size} (\text{bases}) + \text{size}(\text{addressing matrices}) \approx NC\text{size(image)}
\end{equation}
where size$(\cdot)$ is the total size of a tensor, assuming the numbers are stored as floats. For a given number of bases, we calculate the corresponding maximum number of addressing matrices allowed using eqn.~\ref{eqn:budget-eqn} and use the integer lowerbound as the final value.

\begin{table*}[t]
\begin{center}
\scalebox{0.83}{

\begin{tabular}{lcccccccccc}
\toprule
 & I/C & DC~\cite{zhao2021DC} & DSA~\cite{zhao2021DSA} & KIP (NN)~\cite{nguyen2021dataset}  & CAFE$^*$~\cite{wang2022cafe} & TM~\cite{cazenavette2022dataset} & DM~\cite{zhao2021dataset} & Ours \\
\hline
\hline
\multirow{3}{*}{MNIST~\cite{deng2012mnist}} & 1 & 91.7$\pm$0.5 & 88.7$\pm$0.6 & 90.1$\pm$0.1 & 93.1${\pm}$0.3 & - & 89.7$\pm$0.6  & \textbf{98.7$\pm$0.7} \\

& 10 & 97.4$\pm$0.2 & 97.8$\pm$0.1 & 97.5$\pm$0.0 & 97.5${\pm}$0.1 & - & 97.5$\pm$0.1  & \textbf{99.3$\pm$0.5} \\

& 50 & 98.8$\pm$0.2 & 99.2$\pm$0.1 & 98.3$\pm$0.1 & 98.9${\pm}$0.2 & - & 98.6$\pm$0.1 & \textbf{99.4$\pm$0.4} \\
\hline
\multirow{3}{*}{F-MNIST~\cite{xiao2017fashion}} & 1 & 70.5$\pm$0.6 & 70.6${\pm}$0.6 & 73.5${\pm}$0.5 & 77.1${\pm}$0.9 & - & - & \textbf{88.5$\pm$0.1} \\

& 10 &  82.3$\pm$0.4 & 84.6${\pm}$0.3 & 86.8${\pm}$0.1 & 83.0${\pm}$0.3 & - & - & \textbf{90.0$\pm$0.7} \\

& 50 & 83.6$\pm$0.4 & 88.7${\pm}$0.2 & 88.0${\pm}$0.1 & 88.2${\pm}$0.3 & - & - & \textbf{91.2$\pm$0.3} \\
\hline
\multirow{3}{*}{SVHN~\cite{netzer2011reading}} & 1 & 31.2${\pm}$1.4 & 27.5${\pm}$1.4 & 57.3${\pm}$0.1 & 42.9${\pm}$3.0 & - & - & \textbf{87.3${\pm}$0.1} \\

& 10 &  76.1${\pm}$0.6 & 79.2${\pm}$0.5 & 75.0${\pm}$0.1 & 77.9${\pm}$0.6 & - & - & \textbf{89.1${\pm}$0.2} \\

& 50 & 82.3${\pm}$0.3 & 84.4${\pm}$0.4 & 80.5${\pm}$0.1 & 82.3${\pm}$0.4 & - & - & \textbf{89.5$\pm$0.2} \\
\hline
\multirow{3}{*}{CIFAR10~\cite{krizhevsky2009learning}} & 1 & 28.3${\pm}$0.5 & 28.8${\pm}$0.7 & 49.9${\pm}$0.2 & 31.6${\pm}$0.8 & 46.3${\pm}$0.8 & 26.0${\pm}$0.8 & \textbf{66.4$\pm$0.4} \\

& 10 & 44.9${\pm}$0.5 & 52.1${\pm}$0.5 & 62.7${\pm}$0.3 & 50.9${\pm}$0.5  & 65.3${\pm}$0.7& 48.9${\pm}$0.6 & \textbf{71.2$\pm$0.4} \\

& 50 &  53.9${\pm}$0.5 & 60.6${\pm}$0.5 & 68.6${\pm}$0.2 & 62.3${\pm}$0.4 & 71.6${\pm}$0.2 & 63.0${\pm}$0.4 & \textbf{73.6$\pm$0.5} \\
\hline
\multirow{2}{*}{CIFAR100~\cite{krizhevsky2009learning}} & 1 & 12.8${\pm}$0.3 & 13.9${\pm}$0.3 & 15.7${\pm}$0.2 & 14.0${\pm}$0.3 & 24.3${\pm}$0.3 & 11.4${\pm}$0.3 & \textbf{34.0${\pm}$0.4} \\

& 10 & 25.2${\pm}$0.3 & 32.3${\pm}$0.3 & 28.3${\pm}$0.1 & 31.5${\pm}$0.2  & 40.1${\pm}$0.4 & 29.7${\pm}$0.3 & \textbf{42.9${\pm}$0.7} \\
\hline
TinyImageNet~\cite{le2015tiny} & 1 & - & - & - & - & 8.8$\pm$0.3 & 3.9$\pm$0.2 & \textbf{16.0$\pm$0.7} \\
\bottomrule
\end{tabular}

}
\end{center}
\caption{We compare our method with previous works on ConvNet recovered accuracy. Our algorithm consistently outperforms all previous methods and achieves state-of-the-art. $^*$Note that we selected the best results from baseline model variants. I/C is images per class (storage budget eqn.~\ref{eqn:budget-eqn}).}\label{tbl:main}
\vspace{-2mm}

\end{table*}


\begin{wrapfigure}{r}{0.3\textwidth}
\vspace{-6mm}
  \begin{center}
    \includegraphics[width=0.3\textwidth]{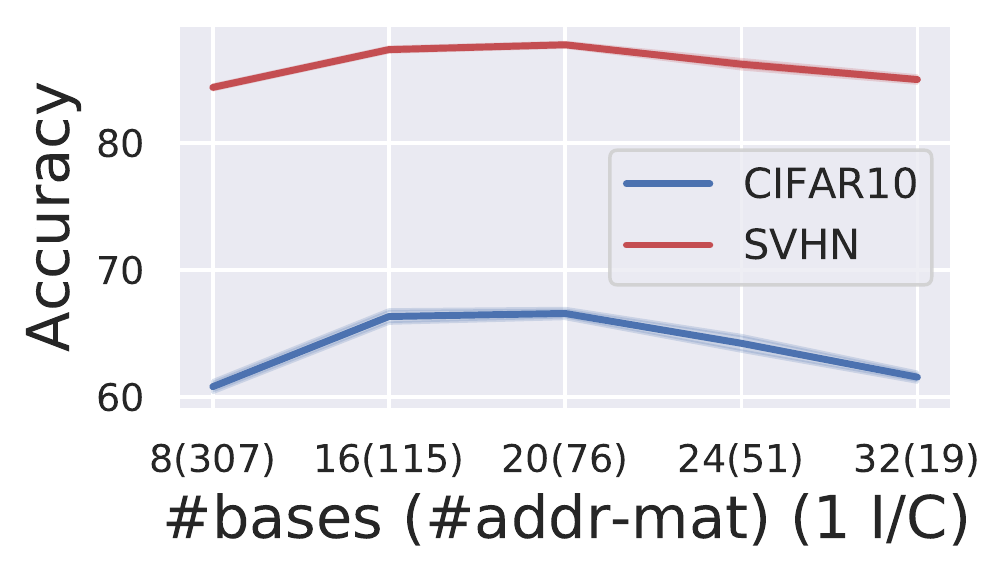}
  \end{center}
  \vspace{-4mm}
  \caption{Validation set}\label{fig:nbasis}
  \vspace{-4mm}
\end{wrapfigure}
\textbf{Model details.} We use bases with spatial resolution downsampled by a factor of $2$ in both height and width from the standard image size based on datasets. All models are trained for 50k iterations with SGD optimizer. For the inner loop optimization, we set the momentum rate as 0.9, and use 150 steps for small memory budgets 1 and 10, and 200 steps for budget 50. The number of bases is selected on the held-out validation set (10\% of training set), an example is shown in figure~\ref{fig:nbasis}, where different numbers of bases and addressing matrices (calculated with eqn.~\ref{eqn:budget-eqn}) have impacts on accuracies; more details  in the appendix. 


\textbf{Result 1: state-of-the-art accuracy.} We compare our model with previous methods: Dataset Condensation (DC)~\cite{zhao2021DC}, Differentiable Siamese Augmentation (DSA)~\cite{zhao2021DSA}, Kernel Inducing Points (KIP)~\cite{nguyen2021dataset}, Distribution Matching (DM)~\cite{zhao2021dataset}, Aligning Features (CAFE)~\cite{wang2022cafe} and Trajectory Matching (TM)~\cite{cazenavette2022dataset}. The results are summarized in table~\ref{tbl:main}. Following previous methods~\cite{zhao2021DSA,nguyen2021dataset,cazenavette2022dataset}, we adopt simple data augmentations and preprocessing: flip and rotation on CIFAR10 and CIFAR100 datasets, and ZCA on SVHN, CIFAR10 and CIFAR100. As shown in table~\ref{tbl:main}, our model consistently outperforms previous methods, especially under high compression rate cases where only 1 image is allowed per class. For example, we achieve 87.3\% and 66.4\% on SVHN and CIFAR10 with 1 image per class, outperforming prior arts by 30\% and 16.5\% under the same storage budget, respectively, and even beat the performance of previous methods using 10 images per class.

\begin{figure*}[tp!]
  \begin{minipage}[t]{0.67\linewidth}
  \begin{subfigure}[t]{0.5\textwidth}
    \centering
    \includegraphics[width=\textwidth]{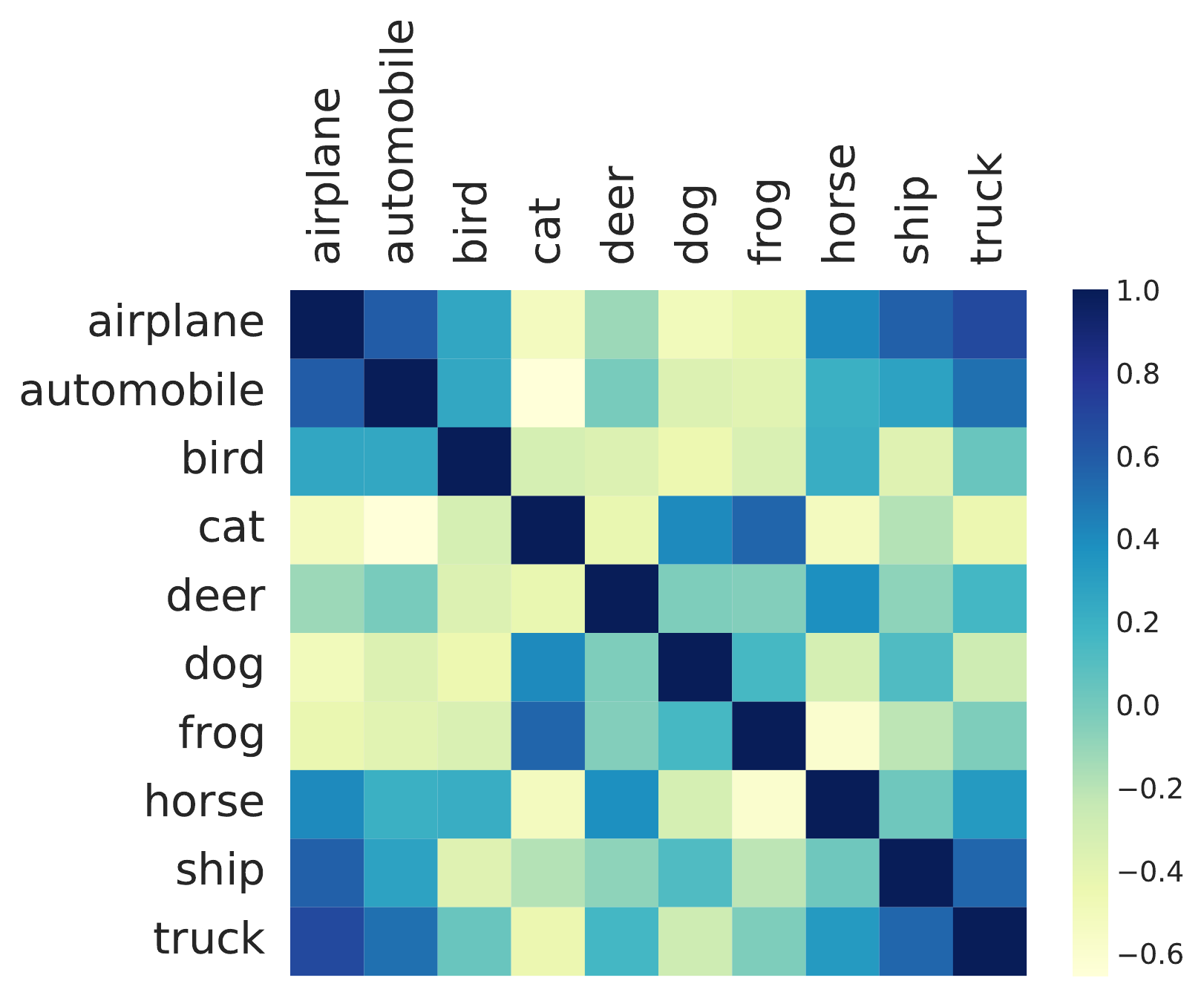}
  \end{subfigure}
  \begin{subfigure}[t]{0.5\textwidth}
    \centering
    \includegraphics[width=\textwidth]{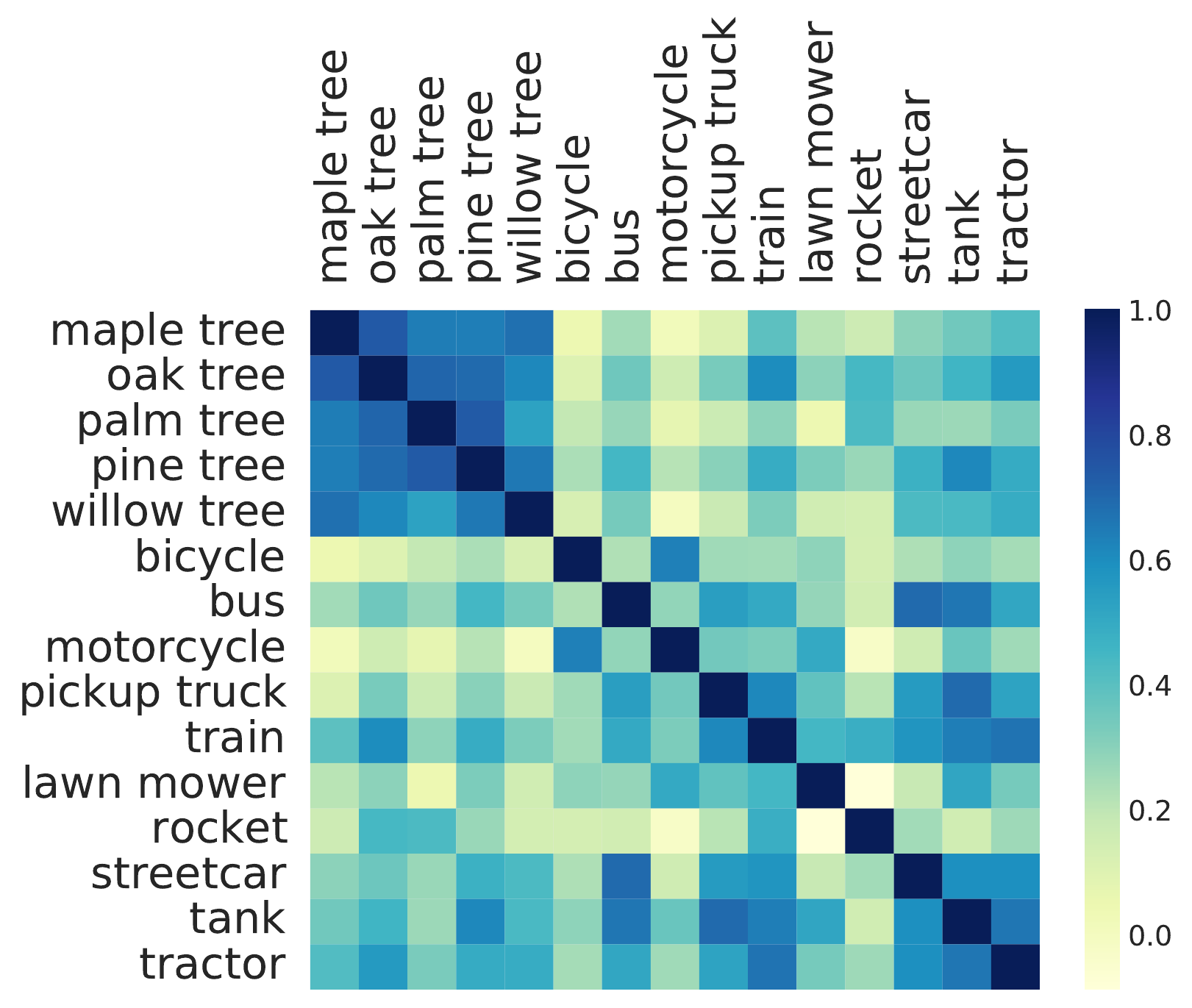}
  \end{subfigure}
  \caption{Similarity matrices of learned addressing coefficients for the CIFAR10 dataset (left) and a subset of CIFAR100 classes (right). 
  }\label{fig:sim-mat-cifar}
  \end{minipage}\hfill
  \begin{minipage}[t]{.28\linewidth}
    \centering
    \includegraphics[width=\textwidth]{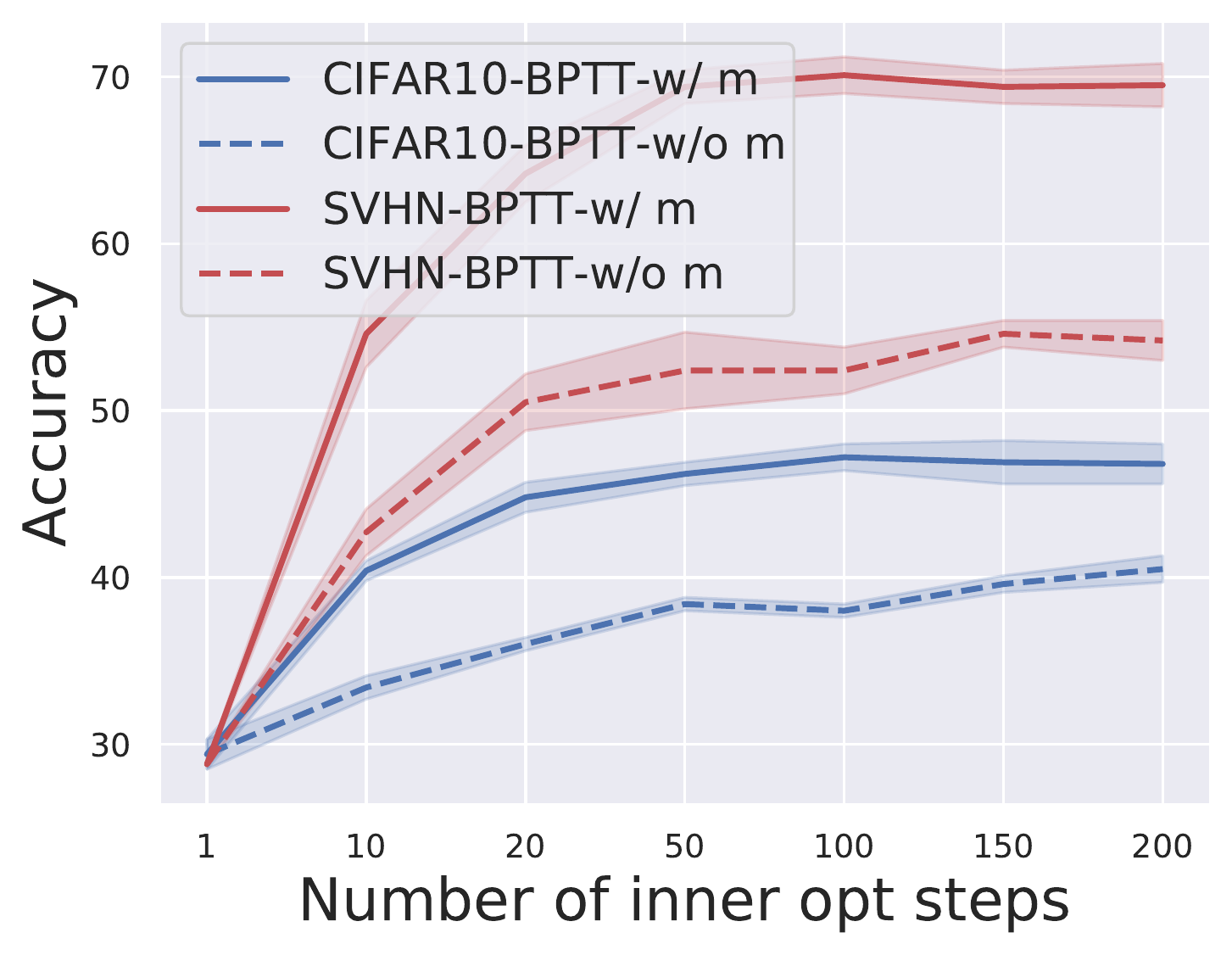}
    \caption{Analysis on BPTT steps and momentums.}\label{fig:bptt}
  \end{minipage}
\end{figure*}

\textbf{Analysis: information sharing across classes.} The core observation in our method is that a common representation can enable information sharing across classes and reduce redundancies. To verify this, we calculate the average coefficients $\Bar{\bv}=\frac{1}{r}\sum_{i=0}^{r-1}\by^T \bA_i$ for each class $\by$ and visualize the cosine similarities of $\Bar{\bv}$ from two classes. The visualizations are shown in fig.~\ref{fig:sim-mat-cifar}. Higher cosine similarity scores indicate that two classes are utilizing similar bases components in the memories to produce synthetic images. For example, in CIFAR100 (right one of figure~\ref{fig:sim-mat-cifar}), classes maple, oak, palm, pine and willow trees have strong sharing, while lawn mower and rocket are distinct from each other. Similar patterns can be found in CIFAR10 dataset, shown in the left one of figure~\ref{fig:sim-mat-cifar}.

\begin{table}[h]
\begin{center}
\small
\begin{tabular}{lcccccc}
\toprule
 & I/C & Single-step GM & Ours$^{\text{BPTT}}$ & Ours$^{\text{BPTT+ds}}$ & Ours$^{\text{Full w/o Aug.}}$ & Ours$^{\text{Full}}$\\
\hline
\hline
 \multirow{3}{*}{CIFAR10} 
 & 1 & 28.8${\pm}$0.7 & 49.1${\pm}$0.6 & 55.2${\pm}$0.5 & 64.2${\pm}$0.6 & \textbf{66.4${\pm}$0.4} \\
 
 & 10 & 52.1${\pm}$0.5 & 62.4${\pm}$0.4 & 65.9${\pm}$0.4 & 70.9${\pm}$0.4 & \textbf{71.2${\pm}$0.4} \\
 
 & 50 & 60.6${\pm}$0.5 & 70.5${\pm}$0.4 & 71.1${\pm}$0.5 & 72.1${\pm}$0.5 & \textbf{73.8${\pm}$0.4} \\

 \hline
 \multirow{2}{*}{CIFAR100}
 
  & 1 & 13.9${\pm}$0.3 & 21.3${\pm}$0.6 & 25.9${\pm}$0.4 & 33.5${\pm}$0.2 & \textbf{34.0${\pm}$0.4}  \\
 
 & 10 & 32.3${\pm}$0.3 & 34.7${\pm}$0.5 & 36.5${\pm}$0.4 & 40.6${\pm}$0.3 & \textbf{42.9${\pm}$0.7} \\
\bottomrule
\end{tabular}
\end{center}
\caption{Ablation studies of every component and comparison with single-step gradient matching~\cite{zhao2021DC}. ds: downsampling. Aug.: data augmentation.}\label{tbl:abl}
\vspace{-5 mm}
\end{table}

\textbf{Result 2: back-propagation through time is a strong baseline.} In figure~\ref{fig:bptt} and table~\ref{tbl:abl}, we show that a vanilla BPTT variant is already a strong baseline which outperforms previous single-step gradient methods~\cite{zhao2021DC} by 40.4\% on SVHN and 20.3\% on SVHN and CIFAR10 under 1 image per class. Note that the performance on SVHN has doubled the accuracy 31.2\% obtained using single-step gradient matching methods~\cite{zhao2021DC}. In the vanilla BPTT variant, no downsampling (ds) or memory addressing formulation is used. We also analyze the effects of \textit{long unrolls} and \textit{momentum terms} on vanilla BPTT in figure~\ref{fig:bptt}. It is observed that on both short inner loops (10 steps) and long ones (100 steps), adding momentum terms can consistently lead to a strong performance boost, e.g. 7.0\% and 9.2\% on CIFAR10. Using longer inner loop trajectories can also increase the recovered accuracy, e.g. 18.2\% and 42.3\% on CIFAR10 and SVHN, respectively, compared to 1 step cases.



\textbf{Ablation studies.} To further analyze the effects of different components in our algorithm, we perform ablation studies on CIFAR10 and CIFAR100. Besides the vanilla BPTT, the ablation results of components (downsampling, augmentation and memory addressing formulation) are summarized in table~\ref{tbl:abl}. We show that downsampling can indeed reduce spatial redundancies (e.g. improve results from 49.1\% to 55.2\% on CIFAR10 with 1 image per class), and memory addressing formulation can further increase the recovered accuracy (from 55.2\% to 64.2\% on CIFAR10 with 1 image per class). It is also shown that our model is quite robust to the ablation of data augmentation, which has a small effect (1-2\%) on the results. The resulting algorithm is a \textit{simple and effective} framework that uses memory addressing formulation and BPTT with long unrolls to distill datasets.

\textbf{Cross-architecture generalization.} Our memories and addressing matrices are also generalizable across various architectures. We test our algorithm on ConvNet, ResNet12 and AlexNet for training and testing. The results are summarized in the appendix, section~\ref{sec:append:DD}, table~\ref{tbl:cross_arch_main}.

\subsection{Continual learning}
\label{sec:cl}

One of the key usages of memories is to prevent forgetting when a model continually learns through tasks. In this section, we evaluate our algorithm on the standard continual learning benchmarks and show that, due to the strong performance, a simple ``compress-then-recall'' method with our model can already rival with previous state-of-the-arts with complex designs.

\textbf{Efficient lifelong learning.} Following~\cite{chaudhry2018efficient}, we work with the problem where all tasks are streamed in mini-batches and learned in a \textit{single pass}. A learner is allowed to be equipped with a small memory buffer. The data samples after seen will not be available unless stored in the buffer. We use a mini-batch size of 10 to stream the data, following previous works~\cite{gupta2020maml, lopez2017gradient}.

\textbf{Evaluation.} The learner's performance after learning on the task stream is commonly evaluated under two metrics: retained accuracy (RA) and backward-transfer and interference (BTI). RA is the average accuracy of the final trained model on all tasks, and BTI measures the performance difference between after it was learned and after the full training process. Note that our algorithm does not perform actual learner training on the data streams and BTI is not applicable.

\textbf{Benchmarks.} We evaluate our method on three tasks widely used in previous Continual Learning works. MNIST Rotations~\cite{lopez2017gradient} contains 20 tasks with 1,000 samples in each. Every task consists of images rotated by a fixed angle from 0 to 180 degrees. MNIST Permutations~\cite{kirkpatrick2017overcoming} has 20 tasks, and each task contains 1,000 images generated through shuffling the image pixels by a fixed permutation. MANY Permutations~\cite{riemer2018learning} is a longer variant with 100 tasks in total and 200 samples in each. Incremental CIFAR-100~\cite{rebuffi2017icarl,lopez2017gradient} splits the CIFAR100 
dataset into 20 5-way classification tasks as the task stream for learning.

\begin{table*}[t]
\begin{center}
\scalebox{0.79}{

\begin{tabular}{lccccccccr}
\toprule
  & \multicolumn{2}{c}{Rotations} & \multicolumn{2}{c}{Permutations}  & \multicolumn{2}{c}{MANY} & \multicolumn{2}{c}{CIFAR-100}\\
 \cmidrule(lr){2-3}
 \cmidrule(lr){4-5}
 \cmidrule(lr){6-7}
 \cmidrule(lr){8-9}
  & RA$\uparrow$ & BTI$\downarrow$ & RA$\uparrow$ & BTI$\downarrow$ & RA$\uparrow$ & BTI$\downarrow$ & RA$\uparrow$ & BTI$\downarrow$ \\
  \hline
 \textsc{Online}  & 53.38$^{\pm1.53}$ & -5.44 & 55.42$^{\pm0.65}$ & -13.76 & 32.62$^{\pm0.43}$ & -19.06 & 32.62$^{\pm0.43}$ & -19.06   \\
 EWC~\cite{kirkpatrick2017overcoming}     &  57.96$^{\pm1.33}$ & -20.42 & 62.32$^{\pm1.34}$ & -13.32 & 33.10$^{\pm0.14}$ & -18.50 & - & - \\
 GEM~\cite{lopez2017gradient}     & 67.38$^{\pm1.75}$ & -18.02 & 55.42$^{\pm1.10}$ & -24.42 & 39.50$^{\pm0.62}$ & -17.50 &48.27$^{\pm1.10}$& -13.7  \\
 MER~\cite{riemer2018learning}     & 77.42$^{\pm0.78}$ & -5.60 & 73.46$^{\pm0.45}$ & -9.96 & 51.00$^{\pm0.54}$ & -13.57 &51.38$^{\pm1.05}$ & -12.83  \\
 La-M~\cite{gupta2020maml} & 77.42$^{\pm0.65}$ & -8.64 & 74.34$^{\pm0.67}$ &-7.60 & 50.43$^{\pm0.21}$& -10.00 & 61.18$^{\pm1.44}$ & -9.00 \\
 sp-La~\cite{von2021learning}  & 77.77$^{\pm0.58}$ & -8.16 & 76.88$^{\pm0.72}$ & -8.39 & 50.81$^{\pm0.79}$ & -13.73 & - & - \\
 \hline
 Ours  & \textbf{80.32$^{\pm\textbf{0.28}}$}  & N/A & \textbf{78.48$^{\pm\textbf{0.76}}$} & N/A & \textbf{74.07$^{\pm\textbf{0.51}}$} & N/A & \textbf{62.58$^{\pm\textbf{1.1}}$} & N/A\\
\bottomrule
\end{tabular}

}
\end{center}
\caption{We show that ``compress-then-recall" is a strong baseline that outperforms previous methods on four continual learning benchmarks. Baseline numbers are from ~\cite{gupta2020maml} or obtained from public official repos.}\label{tbl:cl}
\vspace{-5mm}
\end{table*}

\textbf{Our model.} Based on our distillation method, we adopt a simple framework to perform continual learning: ``compress then recall''. During the training phase, we do not perform learning on neural networks, instead, the dataset of each task is distilled to memories and the paired addressing matrices. During test phase, we simply fetch the corresponding memories and addressing matrices for each task, and train a new model from scratch to perform classification. \textit{Memory buffer designs.} When a new task starts, we use the full remaining memory buffer to store the samples and perform distillation with both buffer samples and streamed samples. After a task ends, the distilled memories and addressing matrices are stored in the buffer, taking $1/T$ of the space, where $T$ is the total number of tasks. Namely, the buffer size keeps shrinking when more compressed representation of tasks is stored. Note that we compare our model with previous methods under the \textit{exact same memory sizes for fair comparisons}. See the appendix for more details on model and memory designs.

\textbf{Results.} We show that this simple method is already a strong baseline that outperforms prior arts on four benchmarks, summarized in table~\ref{tbl:cl}. Our method is compared with: Online, EWC~\cite{kirkpatrick2017overcoming}, GEM~\cite{lopez2017gradient}, MER~\cite{riemer2018learning}, C-MAML~\cite{gupta2020maml}, La-MAML~\cite{gupta2020maml} and Sparse-LaMAML~\cite{von2021learning}. For example, we can obtain a 23\% boost on MNIST MANY benchmark: from 50.81\% to 74.07\%.

We further compare our model with previous works Kernel Continual Learning~\cite{derakhshani2021kernel} and Stable SGD~\cite{mirzadeh2020understanding} following their settings, where each task in MNIST Rotation and MNIST Permutation contains 60,000 samples instead of 1,000 samples. Our model achieves 87.3$^{\pm0.92}$ and 88.3$^{\pm0.58}$ on Permutated MNIST and Rotated MNIST under their setting, outperforming both KCL (85.5$^{\pm0.78}$ and 81.8$^{\pm0.60}$) and Stable SGD (80.1$^{\pm0.51}$ and 70.8$^{\pm0.78}$). Interestingly, our results also are higher than the multitask upperbound (86.5$^{\pm0.21}$ and 87.3$^{\pm0.47}$), potentially due to that there is task interference in joint training, which can be naturally avoided in our method.

\subsection{Synthesizing new classifiers after learning}
\label{sec:syn_cls}
If we want to memorize the past, what is the benefit of storing the compressed representation rather than a trained model? In this section, we show that our compressed representation can enable flexible synthesis of new classifiers after the learning. Specifically, we demonstrate extrapolating between tasks to train new models, and performing memory recall with images instead of labels, showing the generalizability of our framework on other query forms.

\subsubsection{Extrapolating between tasks}
\label{sec:new_task}
In the real world, tasks often do not come together and a learner, therefore, cannot observe all tasks at once. In current machine learning paradigms, when models are separately trained for disjoint tasks, it has difficulty extrapolating between tasks to build new classifiers. This is different from human learning. We show that storing our compressed representation enables a learner to extrapolate and synthesize new classifiers after learning separately on each task. Specifically, we separate CIFAR100 into 20 disjoint 5-way classification tasks as training tasks. For testing, we select classes that are not seen together during training by randomly choosing $k$ tasks and picking $1$ class from each selected task. We use $k=2$ and $k=5$ to construct 2-way and 5-way classification tasks, and sample 1,000 tasks each for evaluation. To train our models, we independently distill the datasets for 20 training tasks into corresponding memories. For each testing task, the class labels are used as queries to recall the synthetic data from the corresponding memories. The recalled data for each label, although not seen together during training, are used for re-training a $k$-way classifier from scratch. We find that the compressed data can indeed train classifiers on new combinations, for example, we can achieve 72.53\%${\pm 8.74}$ on 2-way classification, and 46.54\%${\pm 6.42}$ on 5-way classification, with 1 image per class storage budget. The upperbound with the full real dataset is 92.23\%${\pm 4.76}$ and 82.72\%${\pm 4.29}$.




\subsubsection{Dataset Distillation extension -- recall the past with images}
\label{sec:img_addr}
We extend the standard setting to recall the past with images: when the label information and task scopes are missing, but a few visual observations can be made, we would like to build classifiers based on the visual data. For example, when we see a bear image and a deer image, but cannot recall the exact word or category, can we recall the memories with images and build a classifier? This is possible with our problem formulation, where the forms of queries are not constrained to labels and we can \textit{distill a dataset to memories addressable by images}.

\begin{wraptable}{r}{5cm}
\vspace{-5mm}
\small
\begin{center}
\begin{tabular}{lcc}
\toprule
Methods & 1 shot & 5 shot \\
\hline
 \hline

 Nearest neighbor & 48.55 & 61.72 \\

 classify-then-recall & 50.58 & 58.46 \\
 
 image addressing & \textbf{55.74} & \textbf{71.20} \\
 
\bottomrule
\end{tabular}
\end{center}
\vspace{-3mm}
\caption{Few-shot perf. recovery.}\label{tbl:img_addr}
\vspace{-3mm}
\end{wraptable}



We formulate the problem as follows. Formally, after observing a training dataset $\mathcal{D}_{tr}$ with $\mathcal{Y}=\{0,...,C\textit{-}1\}$, we would like to flexibly build classifiers for a subtask $\mathcal{Y}_g \subset \mathcal{Y}$ based on visual observations $\mathcal{X}_g$ from $\mathcal{Y}_g$, when the actual information of $\mathcal{Y}_g$ is unknown. We work with 1-shot and 5-shot observation cases. As a baseline, we build a nearest neighbor classifier, which is pretrained on $\mathcal{D}_{tr}$ and takes features of few-shot data to classify test images. As a model variant, we could also ``classify-then-recall'', using a classifier trained on $\mathcal{D}_{tr}$ to map the image shots into labels and turning into the standard setup. Benefiting from the general design of our formulation, we show that a model can directly perform ``image addressing'', where a feature network can provide query vectors $\by$ in fig.\ref{fig:main_model}. The feature network, memories and addressing matrices can be jointly trained on $\mathcal{D}_{tr}$. We evaluate the above models and baselines on CIFAR100 and summarize the results in table~\ref{tbl:img_addr}. As shown in the table, our model is not only able to successfully perform the \textit{continuous query addressing} with image feature vectors, but also outperforms two strong baselines on constructing new classifiers. More analysis is in the appendix.

\section{Conclusion and limitations}

In this paper, we propose a framework that distills a large dataset into compact addressable memories. This framework introduces several benefits, including removing the linear growth contraints on the compressed data size, allowing more general queries besides categorical labels, and most importantly, achieving high compression rate with strong re-training performance, outperforming previous state-of-the-arts in dataset distillation. We also demonstrate a ``compress-then-recall'' method using our framework, leading to new state-of-the-arts in continual learning on four datasets. Our full model has potential limitations on the costly inner optimization loop, which might be time-consuming on larger models or datasets. This limitation might be solved by combining the memory formulation with a different learning framework. One potential societal concern with dataset distillation in general is that the distilled dataset may not contain the full diversity of the original data distribution, causing the retrained classifier to perform especially poorly on minority populations; our method arguably takes a step towards mitigating that concern through improving the retrained accuracy.  

\section{Acknowledgements}

This material is based upon work supported by the National Science Foundation under Grants No. 2107048 and 2112562. Any opinions, findings, and conclusions or recommendations expressed in this material are those of the author(s) and do not necessarily reflect the views of the National Science Foundation. We would also like to thank Vishvak Murahari, Sunny Cui, Ruth Fong, Vikram Ramaswamy, and Zeyu Wang for discussions.


\bibliographystyle{unsrt}
\bibliography{neurips_2022}

\begin{thebibliography}{10}

\bibitem{brady2009compression}
Timothy~F Brady, Talia Konkle, and George~A Alvarez.
\newblock Compression in visual working memory: using statistical regularities
  to form more efficient memory representations.
\newblock {\em Journal of Experimental Psychology: General}, 138(4):487, 2009.

\bibitem{loftus2019human}
Geoffrey~R Loftus and Elizabeth~F Loftus.
\newblock {\em Human memory: The processing of information}.
\newblock Psychology Press, 2019.

\bibitem{anderson2014human}
John~R Anderson and Gordon~H Bower.
\newblock {\em Human associative memory}.
\newblock Psychology press, 2014.

\bibitem{goodfellow2013empirical}
Ian~J Goodfellow, Mehdi Mirza, Da~Xiao, Aaron Courville, and Yoshua Bengio.
\newblock An empirical investigation of catastrophic forgetting in
  gradient-based neural networks.
\newblock {\em arXiv preprint arXiv:1312.6211}, 2013.

\bibitem{mccloskey1989catastrophic}
Michael McCloskey and Neal~J Cohen.
\newblock Catastrophic interference in connectionist networks: The sequential
  learning problem.
\newblock In {\em Psychology of learning and motivation}, volume~24, pages
  109--165. Elsevier, 1989.

\bibitem{wang2018dataset}
Tongzhou Wang, Jun-Yan Zhu, Antonio Torralba, and Alexei~A Efros.
\newblock Dataset distillation.
\newblock {\em arXiv preprint arXiv:1811.10959}, 2018.

\bibitem{goodfellow2014generative}
Ian Goodfellow, Jean Pouget-Abadie, Mehdi Mirza, Bing Xu, David Warde-Farley,
  Sherjil Ozair, Aaron Courville, and Yoshua Bengio.
\newblock Generative adversarial nets.
\newblock {\em Advances in neural information processing systems}, 27, 2014.

\bibitem{gemp2020eigengame}
Ian Gemp, Brian McWilliams, Claire Vernade, and Thore Graepel.
\newblock Eigengame: Pca as a nash equilibrium.
\newblock In {\em International Conference on Learning Representations}, 2020.

\bibitem{kingma2013auto}
Diederik~P Kingma and Max Welling.
\newblock Auto-encoding variational bayes.
\newblock {\em arXiv preprint arXiv:1312.6114}, 2013.

\bibitem{zhao2021DC}
Bo~Zhao, Konda~Reddy Mopuri, and Hakan Bilen.
\newblock Dataset condensation with gradient matching.
\newblock In {\em International Conference on Learning Representations}, 2021.

\bibitem{zhao2021DSA}
Bo~Zhao and Hakan Bilen.
\newblock Dataset condensation with differentiable siamese augmentation.
\newblock In {\em International Conference on Machine Learning}, 2021.

\bibitem{nguyen2021dataset}
Timothy Nguyen, Roman Novak, Lechao Xiao, and Jaehoon Lee.
\newblock Dataset distillation with infinitely wide convolutional networks.
\newblock {\em Advances in Neural Information Processing Systems}, 34, 2021.

\bibitem{nguyen2021iclr}
Timothy Nguyen, Zhourong Chen, and Jaehoon Lee.
\newblock Dataset meta-learning from kernel ridge-regression.
\newblock In {\em International Conference on Learning Representations}, 2021.

\bibitem{zhao2021dataset}
Bo~Zhao and Hakan Bilen.
\newblock Dataset condensation with distribution matching.
\newblock {\em arXiv preprint arXiv:2110.04181}, 2021.

\bibitem{cazenavette2022dataset}
George Cazenavette, Tongzhou Wang, Antonio Torralba, Alexei~A Efros, and
  Jun-Yan Zhu.
\newblock Dataset distillation by matching training trajectories.
\newblock In {\em Proceedings of the IEEE/CVF Conference on Computer Vision and
  Pattern Recognition}, pages 4750--4759, 2022.

\bibitem{sucholutsky2019soft}
Ilia Sucholutsky and Matthias Schonlau.
\newblock Soft-label dataset distillation and text dataset distillation.
\newblock {\em arXiv preprint arXiv:1910.02551}, 2019.

\bibitem{riemer2018learning}
Matthew Riemer, Ignacio Cases, Robert Ajemian, Miao Liu, Irina Rish, Yuhai Tu,
  and Gerald Tesauro.
\newblock Learning to learn without forgetting by maximizing transfer and
  minimizing interference.
\newblock {\em arXiv preprint arXiv:1810.11910}, 2018.

\bibitem{finn2017model}
Chelsea Finn, Pieter Abbeel, and Sergey Levine.
\newblock Model-agnostic meta-learning for fast adaptation of deep networks.
\newblock In {\em International Conference on Machine Learning}, pages
  1126--1135. PMLR, 2017.

\bibitem{franceschi2017forward}
Luca Franceschi, Michele Donini, Paolo Frasconi, and Massimiliano Pontil.
\newblock Forward and reverse gradient-based hyperparameter optimization.
\newblock In {\em International Conference on Machine Learning}, pages
  1165--1173. PMLR, 2017.

\bibitem{raghu2020teaching}
Aniruddh Raghu, Maithra Raghu, Simon Kornblith, David Duvenaud, and Geoffrey
  Hinton.
\newblock Teaching with commentaries.
\newblock In {\em International Conference on Learning Representations}, 2020.

\bibitem{reddy2021pragmatic}
Sid Reddy, Anca Dragan, and Sergey Levine.
\newblock Pragmatic image compression for human-in-the-loop decision-making.
\newblock {\em Advances in Neural Information Processing Systems}, 34, 2021.

\bibitem{zhao2021comparing}
Shengjia Zhao, Abhishek Sinha, Yutong He, Aidan Perreault, Jiaming Song, and
  Stefano Ermon.
\newblock Comparing distributions by measuring differences that affect decision
  making.
\newblock In {\em International Conference on Learning Representations}, 2021.

\bibitem{ratcliff1990connectionist}
Roger Ratcliff.
\newblock Connectionist models of recognition memory: constraints imposed by
  learning and forgetting functions.
\newblock {\em Psychological review}, 97(2):285, 1990.

\bibitem{lopez2017gradient}
David Lopez-Paz and Marc'Aurelio Ranzato.
\newblock Gradient episodic memory for continual learning.
\newblock {\em Advances in neural information processing systems},
  30:6467--6476, 2017.

\bibitem{chaudhry2019tiny}
Arslan Chaudhry, Marcus Rohrbach, Mohamed Elhoseiny, Thalaiyasingam Ajanthan,
  Puneet~K Dokania, Philip~HS Torr, and Marc'Aurelio Ranzato.
\newblock On tiny episodic memories in continual learning.
\newblock {\em arXiv preprint arXiv:1902.10486}, 2019.

\bibitem{veniat2020efficient}
Tom Veniat, Ludovic Denoyer, and Marc'Aurelio Ranzato.
\newblock Efficient continual learning with modular networks and task-driven
  priors.
\newblock {\em arXiv preprint arXiv:2012.12631}, 2020.

\bibitem{nguyen2017variational}
Cuong~V Nguyen, Yingzhen Li, Thang~D Bui, and Richard~E Turner.
\newblock Variational continual learning.
\newblock {\em arXiv preprint arXiv:1710.10628}, 2017.

\bibitem{gupta2020maml}
Gunshi Gupta, Karmesh Yadav, and Liam Paull.
\newblock La-maml: Look-ahead meta learning for continual learning.
\newblock {\em arXiv preprint arXiv:2007.13904}, 2020.

\bibitem{rebuffi2017icarl}
Sylvestre-Alvise Rebuffi, Alexander Kolesnikov, Georg Sperl, and Christoph~H
  Lampert.
\newblock icarl: Incremental classifier and representation learning.
\newblock In {\em Proceedings of the IEEE conference on Computer Vision and
  Pattern Recognition}, pages 2001--2010, 2017.

\bibitem{von2021learning}
Johannes Von~Oswald, Dominic Zhao, Seijin Kobayashi, Simon Schug, Massimo
  Caccia, Nicolas Zucchet, and Jo{\~a}o Sacramento.
\newblock Learning where to learn: Gradient sparsity in meta and continual
  learning.
\newblock {\em Advances in Neural Information Processing Systems}, 34, 2021.

\bibitem{prabhu2020gdumb}
Ameya Prabhu, Philip~HS Torr, and Puneet~K Dokania.
\newblock Gdumb: A simple approach that questions our progress in continual
  learning.
\newblock In {\em European conference on computer vision}, pages 524--540.
  Springer, 2020.

\bibitem{de2019continual}
Matthias De~Lange, Rahaf Aljundi, Marc Masana, Sarah Parisot, Xu~Jia, Ales
  Leonardis, Gregory Slabaugh, and Tinne Tuytelaars.
\newblock Continual learning: A comparative study on how to defy forgetting in
  classification tasks.
\newblock {\em arXiv preprint arXiv:1909.08383}, 2(6), 2019.

\bibitem{hadsell2020embracing}
Raia Hadsell, Dushyant Rao, Andrei~A Rusu, and Razvan Pascanu.
\newblock Embracing change: Continual learning in deep neural networks.
\newblock {\em Trends in cognitive sciences}, 24(12):1028--1040, 2020.

\bibitem{rusu2016progressive}
Andrei~A Rusu, Neil~C Rabinowitz, Guillaume Desjardins, Hubert Soyer, James
  Kirkpatrick, Koray Kavukcuoglu, Razvan Pascanu, and Raia Hadsell.
\newblock Progressive neural networks.
\newblock {\em arXiv preprint arXiv:1606.04671}, 2016.

\bibitem{yoon2018lifelong}
Jaehong Yoon, Eunho Yang, Jeongtae Lee, and Sung~Ju Hwang.
\newblock Lifelong learning with dynamically expandable networks.
\newblock In {\em International Conference on Learning Representations}, 2018.

\bibitem{mermillod2013stability}
Martial Mermillod, Aur{\'e}lia Bugaiska, and Patrick Bonin.
\newblock The stability-plasticity dilemma: Investigating the continuum from
  catastrophic forgetting to age-limited learning effects.
\newblock {\em Frontiers in psychology}, 4:504, 2013.

\bibitem{mirzadeh2020understanding}
Seyed~Iman Mirzadeh, Mehrdad Farajtabar, Razvan Pascanu, and Hassan
  Ghasemzadeh.
\newblock Understanding the role of training regimes in continual learning.
\newblock {\em Advances in Neural Information Processing Systems},
  33:7308--7320, 2020.

\bibitem{saha2020gradient}
Gobinda Saha, Isha Garg, and Kaushik Roy.
\newblock Gradient projection memory for continual learning.
\newblock In {\em International Conference on Learning Representations}, 2020.

\bibitem{derakhshani2021kernel}
Mohammad~Mahdi Derakhshani, Xiantong Zhen, Ling Shao, and Cees Snoek.
\newblock Kernel continual learning.
\newblock In {\em International Conference on Machine Learning}, pages
  2621--2631. PMLR, 2021.

\bibitem{nichol2018reptile}
Alex Nichol and John Schulman.
\newblock Reptile: a scalable metalearning algorithm.
\newblock {\em arXiv preprint arXiv:1803.02999}, 2(3):4, 2018.

\bibitem{deng2012mnist}
Li~Deng.
\newblock The mnist database of handwritten digit images for machine learning
  research.
\newblock {\em IEEE Signal Processing Magazine}, 29(6):141--142, 2012.

\bibitem{xiao2017fashion}
Han Xiao, Kashif Rasul, and Roland Vollgraf.
\newblock Fashion-mnist: a novel image dataset for benchmarking machine
  learning algorithms.
\newblock {\em arXiv preprint arXiv:1708.07747}, 2017.

\bibitem{netzer2011reading}
Yuval Netzer, Tao Wang, Adam Coates, Alessandro Bissacco, Bo~Wu, and Andrew~Y
  Ng.
\newblock Reading digits in natural images with unsupervised feature learning.
\newblock 2011.

\bibitem{krizhevsky2009learning}
Alex Krizhevsky, Geoffrey Hinton, et~al.
\newblock Learning multiple layers of features from tiny images.
\newblock 2009.

\bibitem{le2015tiny}
Ya~Le and Xuan Yang.
\newblock Tiny imagenet visual recognition challenge.
\newblock {\em CS 231N}, 7(7):3, 2015.

\bibitem{wang2022cafe}
Kai Wang, Bo~Zhao, Xiangyu Peng, Zheng Zhu, Shuo Yang, Shuo Wang, Guan Huang,
  Hakan Bilen, Xinchao Wang, and Yang You.
\newblock Cafe learning to condense dataset by aligning features.
\newblock In {\em IEEE/CVF Conference on Computer Vision and Pattern
  Recognition 2022}, 2022.

\bibitem{chaudhry2018efficient}
Arslan Chaudhry, Marc'Aurelio Ranzato, Marcus Rohrbach, and Mohamed Elhoseiny.
\newblock Efficient lifelong learning with a-gem.
\newblock {\em arXiv preprint arXiv:1812.00420}, 2018.

\bibitem{kirkpatrick2017overcoming}
James Kirkpatrick, Razvan Pascanu, Neil Rabinowitz, Joel Veness, Guillaume
  Desjardins, Andrei~A Rusu, Kieran Milan, John Quan, Tiago Ramalho, Agnieszka
  Grabska-Barwinska, et~al.
\newblock Overcoming catastrophic forgetting in neural networks.
\newblock {\em Proceedings of the national academy of sciences},
  114(13):3521--3526, 2017.

\end{thebibliography}

\newpage

\appendix

\section{Experiment setups}
In this section, we provide detailed experimental setups for all the tasks discussed in the main paper. Specifically, we will explain the datasets, architectures and implementation details for all tasks.

\subsection{Dataset Distillation}
\label{sec:dd_appendix}
\textbf{Datasets.} Our models are tested on six standard dataset distillation benchmarks: 
\begin{itemize}
    \item MNIST contains 10 classes with 60,000 writing digit images as the training set and 10,000 images as the test set. The images are gray-scale with a shape of $28\times 28$ and associated with a label from 10 classes (digit 0-9). 
    
    \item FashionMNIST is a dataset with clothing and shoe images and consists of a training set with size 60,000 and a test set with size 10,000. Each image is $28\times 28$ in gray scale, and has a label from 10 classes.
    
    \item SVHN street digit images where each image has a shape of $32\times 32 \times 3$. The dataset contains 73257 images for training and 26032 images for testing. We use the cropped SVHN where the center of the image indicates the number and the rest is background. Each image is categorized into 10 classes (digits 0-9).
    
    \item CIFAR10 is a dataset consisting of $32\times 32$ RGB images and has 10 classes in total: airplane, automobile, bird, cat, deer, dog, frog, horse, ship, and truck. Each class contains 5,000 images for training and 1,000 images for testing, leading to 50,000 images for training and 10,000 images for testing in total.
    
    \item CIFAR100 contains 60,000 images in total from 100 classes. For every class, 500 images are used for training and 100 images are used in testing. The 100 classes are associated with 20 superclasses, where each superclass contains 5 classes at a finer level.
    
    \item TinyImageNet is a downscaled subset of ImageNet, with 200 classes. The dataset contains images of shape 64x64, a training set with 100,000 images and a testing set with 10,000 images.
\end{itemize}

\textbf{Architectures.} We mainly work with a three-layer convolutional neural network, denoted as "ConvNet", which contains convolutional layers with $3\times 3$ filters, followed by ReLU activation function and  InstanceNorm. The network has 128 hidden dimensions and uses an average pooling layer with $2\times 2$ kernel size after every Instancenorm operation. We also test our models on ResNet-12 with 64, 128, 256, 512 hidden dimensions in each block. The ResNet-12 architecture is slightly modified by replacing BatchNorm with InstanceNorm, and removing the final average pooling layer. We find using the full spatial information in the final layer is important for distillation. Both ConvNet and ResNet-12 are standard architectures for few-shot learning benchmarks. 

\textbf{Implementation details.} We use one 24-GB GPU for each experiment run. For all our models, we use a SGD optimizer with learning rate 0.1 and momentum rate 0.5. Every model is trained for 50,000 iterations. For both the inner loop optimization and evaluation, we use learning rate 0.01 and momentum rate 0.9. For random initialization of addressing matrices and bases, we use Kaiming uniform initialization. To select the number of bases for each setting, we randomly sample 10\% of training set as the validation set. Data augmentations with rotation and flip are applied on CIFAR10 and CIFAR100 datasets. ZCA preprocessing is used on CIFAR10, CIFAR100 and SVHN datasets. No ZCA preprocessing or data augmentations are used on MNIST and FashionMNIST datasets.

\subsection{Continual learning}

\begin{table}[t]
\begin{center}
\begin{tabular}{lccccccc}
\toprule
 & Dataset & \#Tasks & Batch size & \#Samples/task & Total mem size & \#Bases & \#Replay\\
\hline
\hline
 & Rotations & 20 & 10 & 1000 & 200 & 24(ds) & 4\\
 
 & Permutations & 20 & 10 & 1000 & 200 & 8 & 20 \\
 
 & MANY & 100 & 10 & 1000 & 1000  & 8 & 20\\
 
 & CIFAR100 & 20 & 10 & 2250 & 200  & 24(ds) & 4\\

 & Rotations$^*$ & 20 & 10 & 60000 & 200  & 24(ds) & 2\\
 
 & Permutations$^*$ & 20 & 10 & 60000 & 200 & 8 & 2 \\
 
\bottomrule
\end{tabular}
\end{center}
\caption{The details on six benchmarks used in the experiments: MNIST Rotations (Rotations), MNIST Permutations (Permutations), MANY Permutations (MANY), Incremental CIFAR100 (CIFAR100), MNIST Rotations with 60,000 data samples (Rotations$^*$), MNIST Permutations with 60,000 data samples (Permutations$^*$). Note that works compare under different benchmarks, we follow the settings and compare our model with La-MAML on Rotations, Permutations, MANY, and CIFAR100, and compare with Kernel Continual Learning on Rotations$^*$ and Permutations$^*$. \#Samples per task is specified for training. (ds) indicates using downsampled bases.}\label{tbl:dataset_appendix}
\end{table}

\textbf{Datasets.} We use six datasets to evaluate our models. The details are summarized in table~\ref{tbl:dataset_appendix}.

\textbf{Architectures and implementation details.} For all MNIST-based datasets, we use a multi-layer perceptron (MLP) with 256 hidden units. Following La-MAML, we use the ConvNet architecture with 160 hidden dimensions. All experiments are run on a 24-GB GPU, using a SGD optimizer with 0.1 learning rate and 0.5 momentum rate. The inner loop optimization learning rate is set as 0.01 with momentum rate 0.9. During the testing phase, the re-training phase uses the same setups as the inner loop optimization. We use data samples stored in the memory buffer for minibatch replay to perform compressing, summarized in table~\ref{tbl:dataset_appendix}.

\subsection{New classifier synthesis} 

\textbf{Datasets and setups.} For experiments on both extrapolating between tasks and recall with images, we use CIFAR100 as the dataset. \textit{For task extrapolation experiments}, we split CIFAR100 into 20 5-way classification tasks for training, and use 2-way and 5-way classification for testing. The 2-way and 5-way tasks during testing are obtained through randomly selecting 2 or 5 training tasks and then randomly sampling 1 class from each selected task. This ensures that every pairs of classes in a testing task have not been used together for training. \textit{For recall with images experiments}, we use all classes in CIFAR100 for training, and use 20 5-way classification tasks in testing. During evaluation on a 5-way classification task, we sample 1 or 5 images per class (depends on 1-shot or 5-shot), and use the sampled images for recall. The sampled images are from test set, i.e. we would like to use testing images to perform recall and build a new classifier. 

\textbf{Architectures and implementation details.} In the task extrapolation experiments, since our models are performing dataset distillation, we use the exact same hyperparameters and architectures as dataset distillation tasks in Sec.~\ref{sec:dd_appendix}. For recall with images, we use 64 bases and 16 addressing matrices (i.e. each query can generate 16 synthetic images) in our model. For baselines, we pretrain the feature backbone for nearest neighbor classifier and the classifier in "classify-then-recall" for 100 epochs on CIFAR100, using SGD optimizer with 0.01 learning rate and 0.9 momentum rate. The visual observations (image shots) we used are from the test set.

\section{Additional results and discussion}
\subsection{Dataset Distillation}
\label{sec:append:DD}
\textbf{Back-propagation through time as a strong baseline.} Besides the main ablation study results on CIFAR10 and CIFAR100, table~\ref{tbl:full_ablation} provides the results for the benchmarks. As shown in the table, back-propagation through time is indeed a strong baseline that consistently outperforms the single-step gradient matching method, and downsampling can reduce spatial redundancies and improve the compression rate, leading to a higher recovery performance. 

\begin{table}[h]
\begin{center}
\begin{tabular}{lccccccr}
\toprule 
  & \multicolumn{3}{c}{1 image/class} & \multicolumn{3}{c}{10 images/class}\\
 \cmidrule(lr){2-4}
 \cmidrule(lr){5-7}
 & AlexNet & ResNet-12 & ConvNet & AlexNet & ResNet-12 & ConvNet \\
 \hline
AlexNet  & 58.5$\pm$0.5 & 53.6$\pm$0.6 & 57.3$\pm$0.6 & 65.6$\pm$0.5 & 60.2$\pm$0.6 & 63.7$\pm$0.6 & \\
ResNet12 & 53.2$\pm$0.8 & 58.5$\pm$0.5 & 57.0$\pm$0.3 & 62.3$\pm$0.9 & 67.8$\pm$0.3 & 65.2$\pm$0.6 \\
ConvNet  & 50.5$\pm$1.3 & 55.9$\pm$0.6 & 66.4$\pm$0.4 & 63.8$\pm$0.8 & 67.5$\pm$0.4 & 71.2$\pm$0.4 \\
 
\bottomrule
\end{tabular}
\end{center}
\caption{Cross architecture generalization under various pixel/image storage budgets.}\label{tbl:cross_arch_main}
\end{table}

\begin{table*}[t]
\begin{center}
\small
\begin{tabular}{lcccccccc}
\toprule
& I/C & Single-step GM & Ours$^{\text{BPTT}}$ & Ours$^{\text{BPTT+ds}}$ & Ours$^{\text{Full w/o Aug.}}$ & Ours$^{\text{Full}}$\\
\hline
\hline
\multirow{3}{*}{MNIST} & 1 & 91.7$\pm$0.5 & 95.2$\pm$0.3 & 98.2$\pm$0.1 & - & \textbf{98.7$\pm$0.7} \\

& 10 & 97.4$\pm$0.2 & 98.8$\pm$0.1 & 98.9$\pm$0.1 & - & \textbf{99.3$\pm$0.5} \\

& 50 & 98.8$\pm$0.2 & 99.2$\pm$0.1 & 99.4$\pm$0.1 & - & \textbf{99.4$\pm$0.4} \\
\hline
\multirow{3}{*}{F-MNIST} & 1 & 70.5$\pm$0.6 & 83.9$\pm$0.4 & 86.7$\pm$0.3 & - & \textbf{88.5$\pm$0.1} \\

& 10 &  82.3$\pm$0.4 & 89.1$\pm$0.2 & 89.1$\pm$0.1 & - &  \textbf{90.0$\pm$0.7} \\

& 50 & 83.6$\pm$0.4 & 90.4$\pm$0.1 & 90.7$\pm$0.1 & - & \textbf{91.2$\pm$0.3} \\
\hline
\multirow{3}{*}{SVHN} & 1 & 31.2${\pm}$1.4 & 71.6$\pm$0.8 & 80.1$\pm$0.5& - & \textbf{87.3${\pm}$0.1} \\

& 10 & 76.1${\pm}$0.6 & 83.1$\pm$0.3 & 86.2$\pm$0.2 & - & \textbf{89.1${\pm}$0.2} \\

& 50 & 82.3${\pm}$0.3 & 86.5$\pm$0.2 & 88.8$\pm$0.2 & - & \textbf{89.5$\pm$0.2} \\
\hline
\multirow{3}{*}{CIFAR10} & 1 & 28.3${\pm}$0.5 & 49.1$\pm$0.6 & 55.2$\pm$0.5 & 64.2$\pm$0.6 & \textbf{66.4$\pm$0.4} \\

& 10 & 44.9${\pm}$0.5 & 62.4$\pm$0.4 & 65.9$\pm$0.4 & 70.9$\pm$0.4 & \textbf{71.2$\pm$0.4} \\

& 50 &  53.9${\pm}$0.5 & 70.5$\pm$0.4 & 71.1$\pm$0.5 & 72.1$\pm$0.5 & \textbf{73.6$\pm$0.5} \\
\hline
\multirow{2}{*}{CIFAR100} & 1 & 12.8${\pm}$0.3 & 21.3$\pm$0.6 & 25.9$\pm$0.4 & 33.5$\pm$0.2 & \textbf{34.0${\pm}$0.4} \\

& 10 & 25.2${\pm}$0.3 & 34.7$\pm$0.5 & 36.5$\pm$0.4 &  40.6$\pm$0.3 & \textbf{42.9${\pm}$0.7} \\
\bottomrule
\end{tabular}
\end{center}
\caption{Full ablation studies on model variants and comparison with single-step gradient matching baseline. 
No augmentations are used on MNIST, FashionMNIST and SVHN.}\label{tbl:full_ablation}
\end{table*}

\textbf{Transfer across architectures.} To show that our compressed memories are generalizable across architectures, we also test the training on ResNet-12. Specifically, we learn the memories and addressing matrices on ConvNet and ResNet-12, and test them on ResNet-12 and ConvNet, respectively. Results are summarized in table~\ref{tbl:cross_arch_main}. We use 10 images per class as the storage budget on CIFAR10. Each row is the architecture that our method trains on, and each column is the generalization performance. The learned compressed representation is quite robust across ConvNet and ResNet-12.

\begin{figure}[h]
  \centering
  \includegraphics[width=1\linewidth]{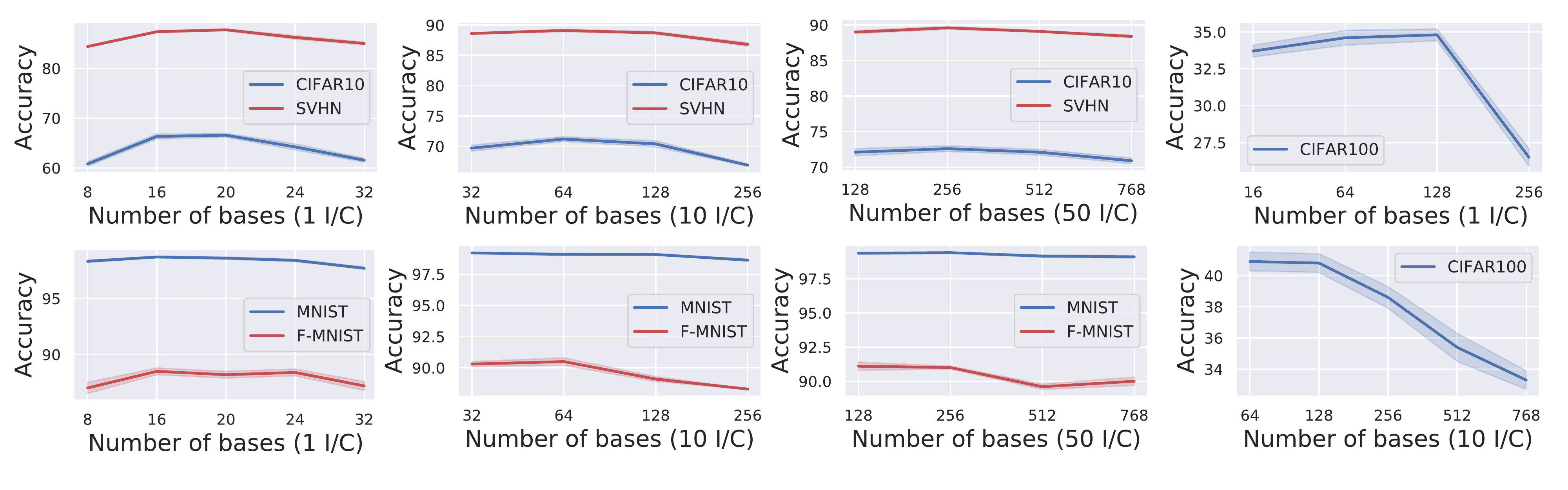}
   \caption{Number of bases v.s. retrain accuracy on validation set. I/C: images per class.}
   \label{fig:full_bases_curves}
\end{figure}

\textbf{Choice of \# bases.} To select the number of bases for each experiment, we evaluate the performance on a separate validation set, which is 10\% random samples of the training set. The results on the validation set are shown in fig.~\ref{fig:full_bases_curves}. We select the number of bases that leads to the highest performance on the validation set for the full training set distillation.

\textbf{Further ablations on momentum terms.} How is the momentum term exactly affecting the backpropagation through time process? We analyze the performance of baseline BPTT algorithms on three cases: no-momentum, forward-only momentum, and full momentum. No-momentum uses BPTT without momentum terms. Forward-only momentum uses the momentum term only in the forward BPTT, but blocks the gradients on the momentum term in the backward pass (except for the gradients on the current time step weights) to remove the ``bridging effect" of momentum term across multiple steps. Full momentum is our full model. All the experiments are performed on CIFAR10 with 200 inner optimization steps.

\begin{table}[h]
\small
\begin{center}
\begin{tabular}{lcccc}
\toprule
 I/C & no-momentum & forward-only momentum & full momentum \\
 \hline
 1 & 40.5$\pm0.8$ & 45.6$\pm0.7$ & 49.1$\pm0.6$ \\
 
 10 & 50.0$\pm0.5$ & 57.4$\pm0.3$ & 62.4$\pm0.4$ \\
 \bottomrule
\end{tabular}
\end{center}
\caption{Further analysis of momentum terms of BPTT on CIFAR10 dataset.}\label{tbl:momentum_appendix}
\vspace{-3mm}
\end{table}

\textbf{Adam optimizer for inner loop.} We also experimented with using Adam optimizer to optimize the synthetic data, instead of using stochastic gradient descent with momentum. Empirically, we found that Adam optimizer leads to certain instability of gradients (e.g., magnitude) on the inner optimization steps when using the same learning rate magnitude and perfers smaller ones such as 1e-4. The end results are similar to the SGD algorithms.


\subsection{Continual learning}
\textbf{Memory designs in ``compress-then-recall''.} We follow the Reservoir sampling strategy to store samples in the memory buffer. When learning through the tasks, our algorithm utilizes all the currently available memory buffer storage space to store samples. After the learning on one task is finished, the algorithm saves the compressed representation to the memory buffer, taking $1/T$ the buffer where $T$ is the total number of tasks, and clear the storage space which stores the real samples for the current task. This strategy makes sure that the compression algorithm has enough samples to replay, resulting in $1 - (t-1) / T$ of the storage to use, where $t \in \{1,...,T\}$ is the current task index.

\subsection{New classifier synthesis}

\textbf{Designs of ``image addressing'' model.} Since our formulation allows flexible query forms, we use an extra ConvNet to take the visual observations (images) as input and treat the output feature vectors as queries. The feature vector queries are used for vector matrix product with addressing matrices to compute coefficients for combining bases. To train the addressing model: For every training iteration, we randomly subsample a subset of classes from all classes and pick 1 or 5 images (depends on 1-shot or 5-shot), and use the recalled synthetic datasets with the feature vectors of the images to perform inner loop optimizations. The generalization loss is computed using other image-label pairs from the subset classes (the same as standard dataset distillation training). The ConvNet (feature extractor), bases and image matrices are jointly trained. 

\textbf{Discussion of ``image addressing'' results.} We compare the "image addressing" model with two strong baselines: nearest neighbor classifiers and "classify-then-recall" method. It's interesting to see that, having the ability to access the dataset-level information (even compressed) can often lead to better performance when building a new classifier, while nearest neighbor classifiers can only utilize image shots to serve as limited information for classification. Note that the "classify-then-recall" method is also a strong baseline, but can suffer from the classification errors on test images, leading to less robust recall. The direct usage of feature vectors from image shots can provide a continuous space and potentially lead to more robust behaviours in the addressing and recall processes.

\section{Visualization and analysis}
\begin{figure}[h]
  \centering
  \includegraphics[width=1.1\linewidth]{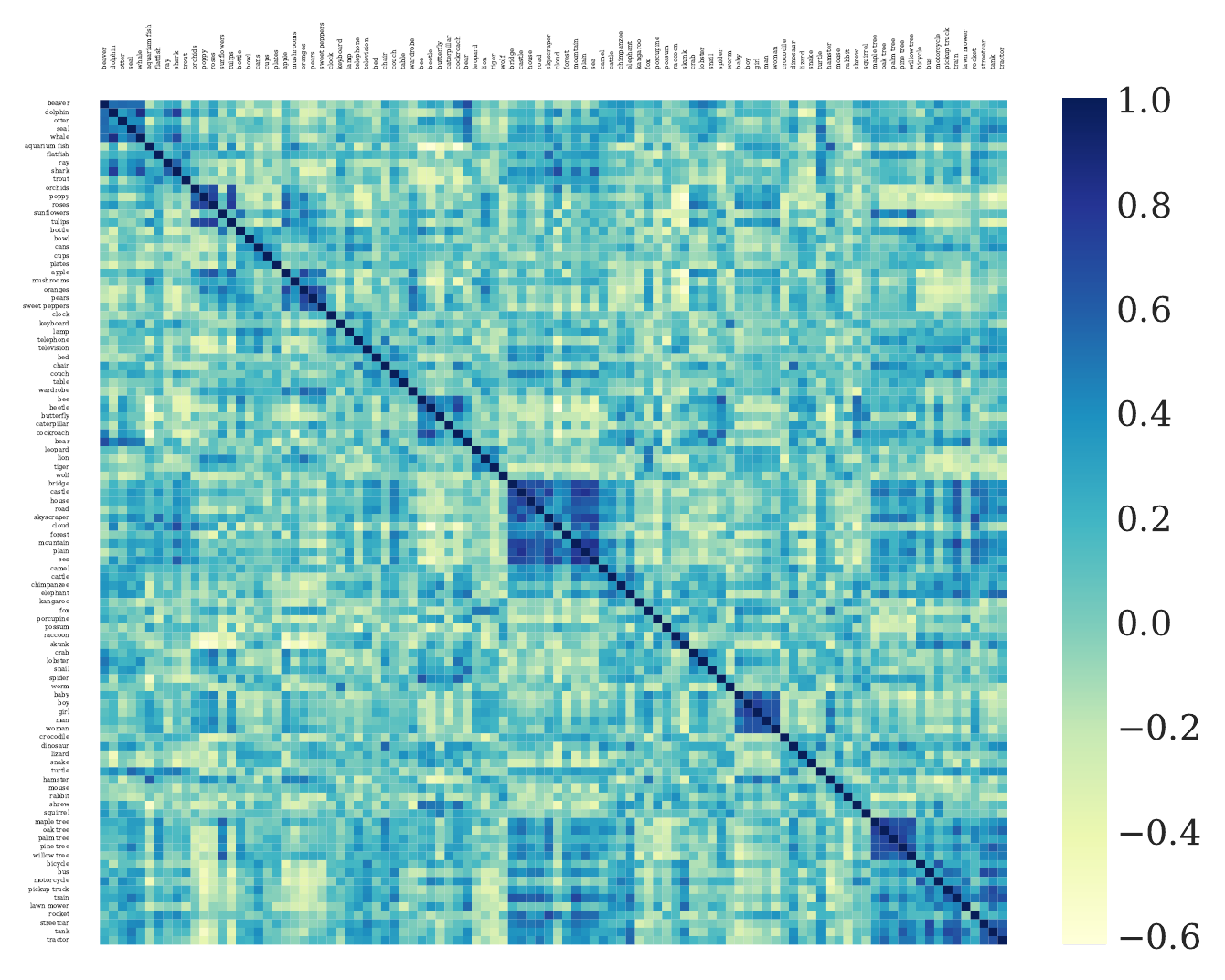}
   \caption{Full coefficient cosine similarity matrix on CIFAR100. Zoom in to view the details. Classes are ordered with superclasses. On the x and y axis, in order, every 5 classes belongs to a common superclass.}
   \label{fig:full_sim_mat}
\end{figure}

\textbf{Coefficients similarity map.} We show the full matrix of cosine similarities on the coefficients that combine the bases from all 100 classes in CIFAR100, as shown in fig.~\ref{fig:full_sim_mat}. The order of classes on x and y axis is organized by superclasses. Every 5 classes is under a common superclass on the axis. As shown by the matrix, we can clearly see that the classes under the same superclass often have significant similarities, indicating strong sharings when combining bases. For example, categories bridge, castle and house share similar bases; baby, girl, man and woman also share similar bases, while crab and tulips use very different coefficients to perform addressing.

\begin{figure}[h]
  \centering
  \includegraphics[width=0.4\linewidth]{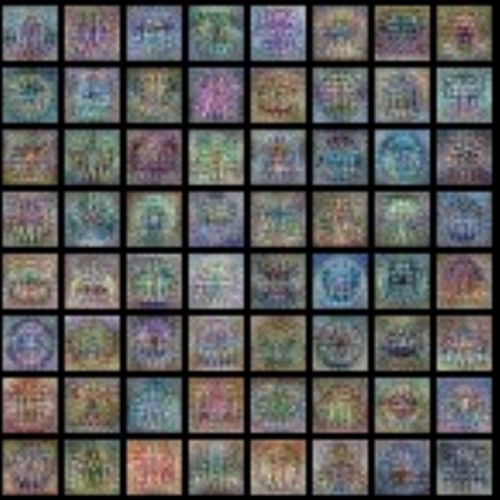}
   \caption{CIFAR100 learned 64 bases.}
   \label{fig:cifar100-bases}
\end{figure}

\textbf{Visualization on bases.} In figure~\ref{fig:cifar100-bases}, we visualize the learned 64 bases on CIFAR100. The bases contain various colors, shapes and textures, and are used to be combined with coefficients generated from queries and addressing matrices.


\section{More visualization comparisons}
In this section, we further compare the visualization of our methods under various settings.

\subsection{Same amount of generated images}
\label{sec:vis-same-amount}
We visualize the synthetic images from the baseline method BPTT and from our proposed memory addressing parameterization. For BPTT, we use 100 image per class as the budgets, and for our method, we use 10 images per class and 43 bases to generate approximately the same amount of recalled images (99). The synthetic images are visualized in figures below. To easily compare with the vanilla version of BPTT, we do not use downsampling in either BPTT or the memory addressing formulation.

\begin{figure}[h]
  \centering
  \includegraphics[width=1\linewidth]{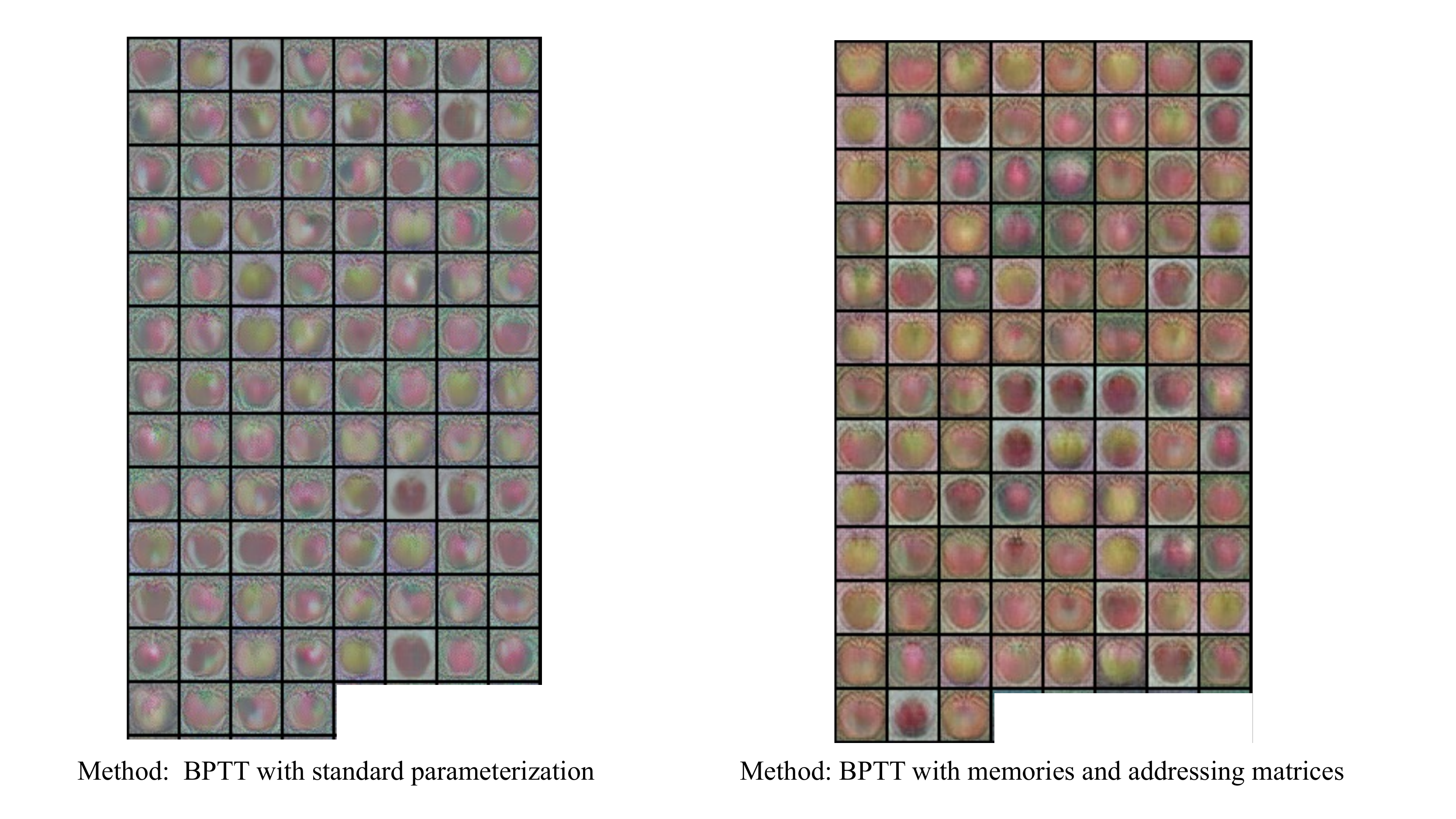}
   \caption{Recalled synthetic images for class apple.}
   \label{fig:vis_apple}
\end{figure}

\begin{figure}[h]
  \centering
  \includegraphics[width=1\linewidth]{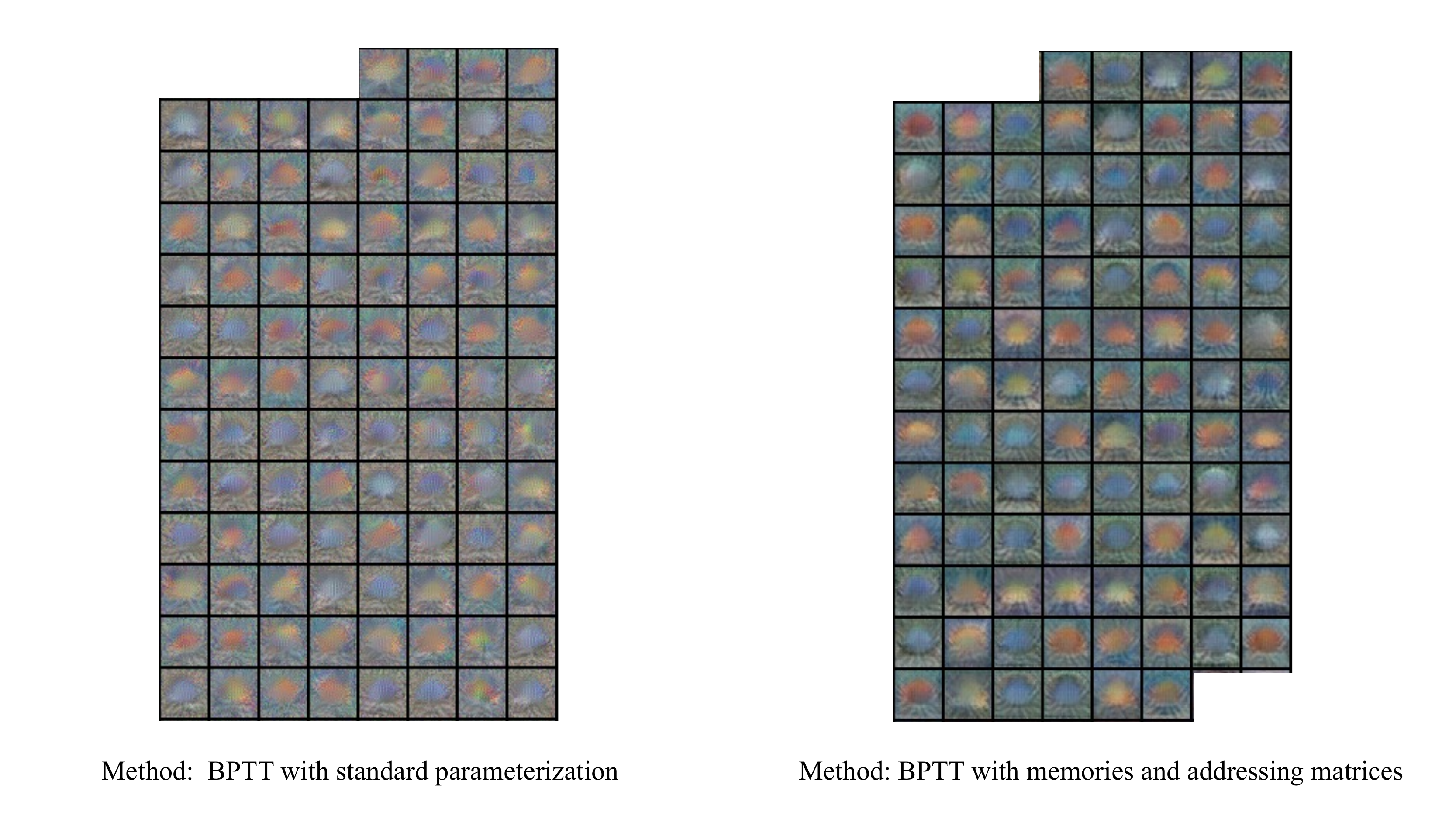}
   \caption{Recalled synthetic images for class aquarium fish.}
   \label{fig:vis_aqua}
\end{figure}


\begin{figure}[h]
  \centering
  \includegraphics[width=1\linewidth]{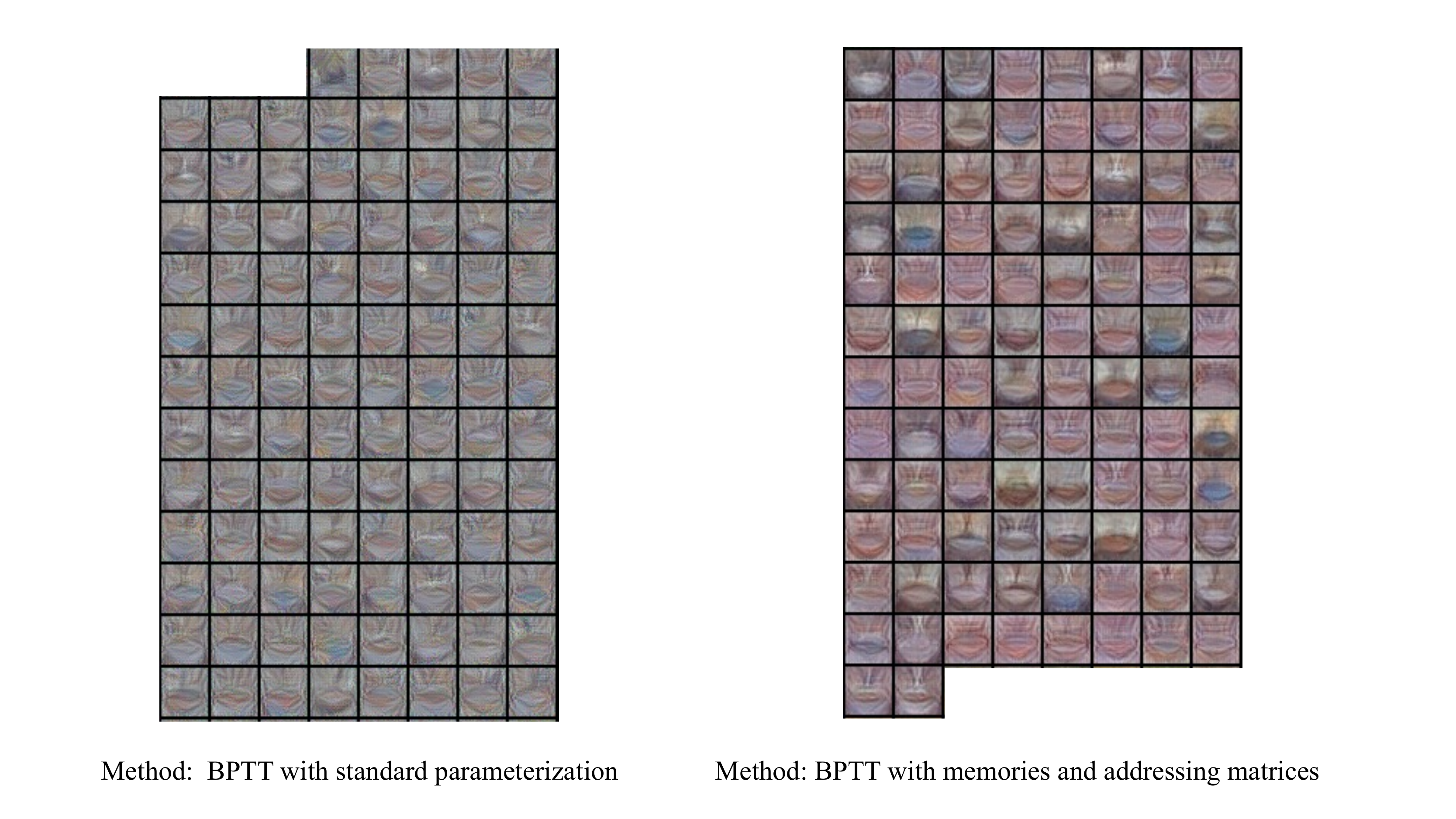}
   \caption{Recalled synthetic images for class bed.}
   \label{fig:vis_bed}
\end{figure}

\begin{figure}[h]
  \centering
  \includegraphics[width=1\linewidth]{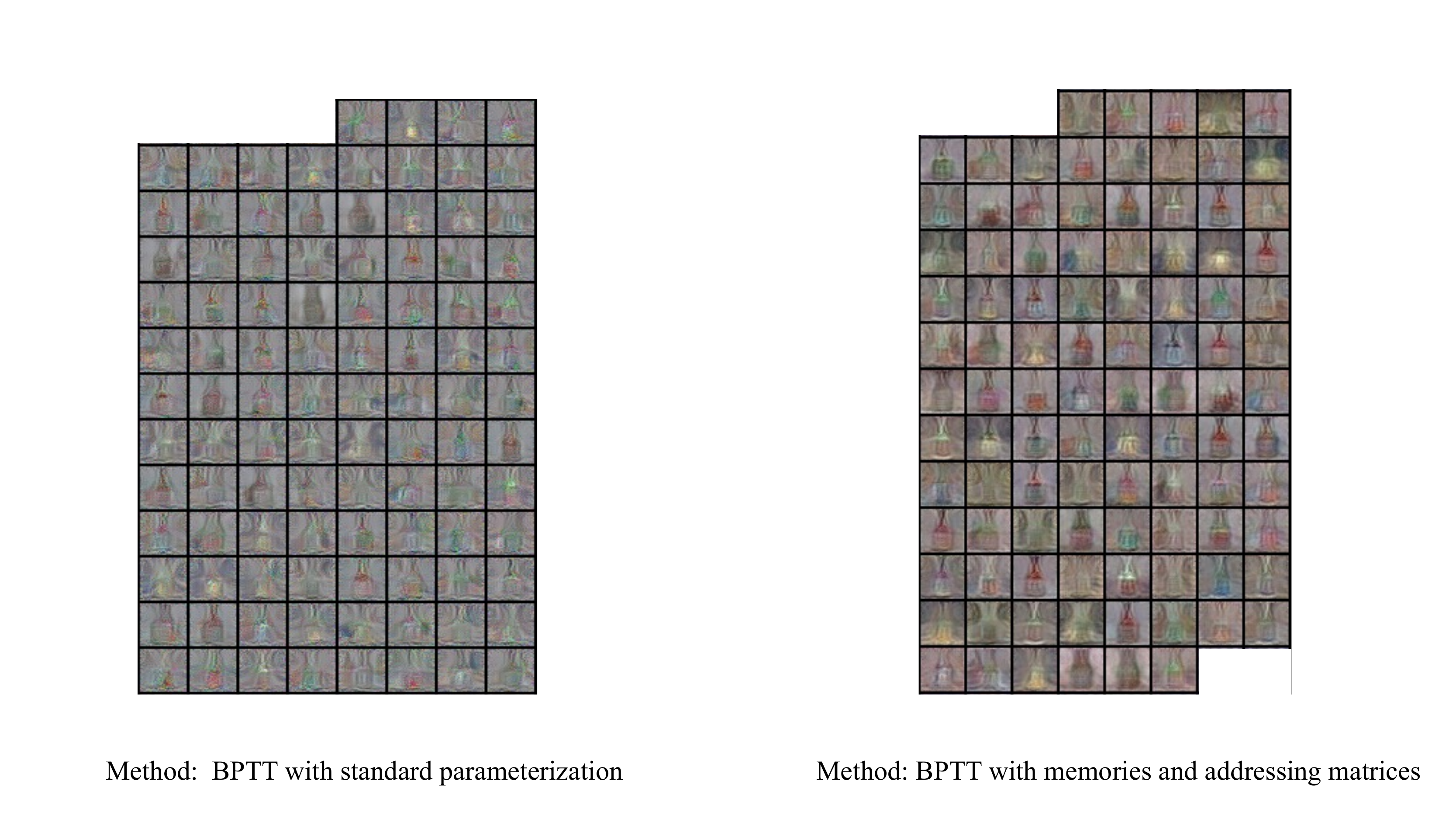}
   \caption{Recalled synthetic images for class bottle.}
   \label{fig:vis_bottle}
\end{figure}

\newpage

\subsection{Various image per class budgets}

\begin{figure}[h]
  \centering
  \includegraphics[width=0.4\linewidth]{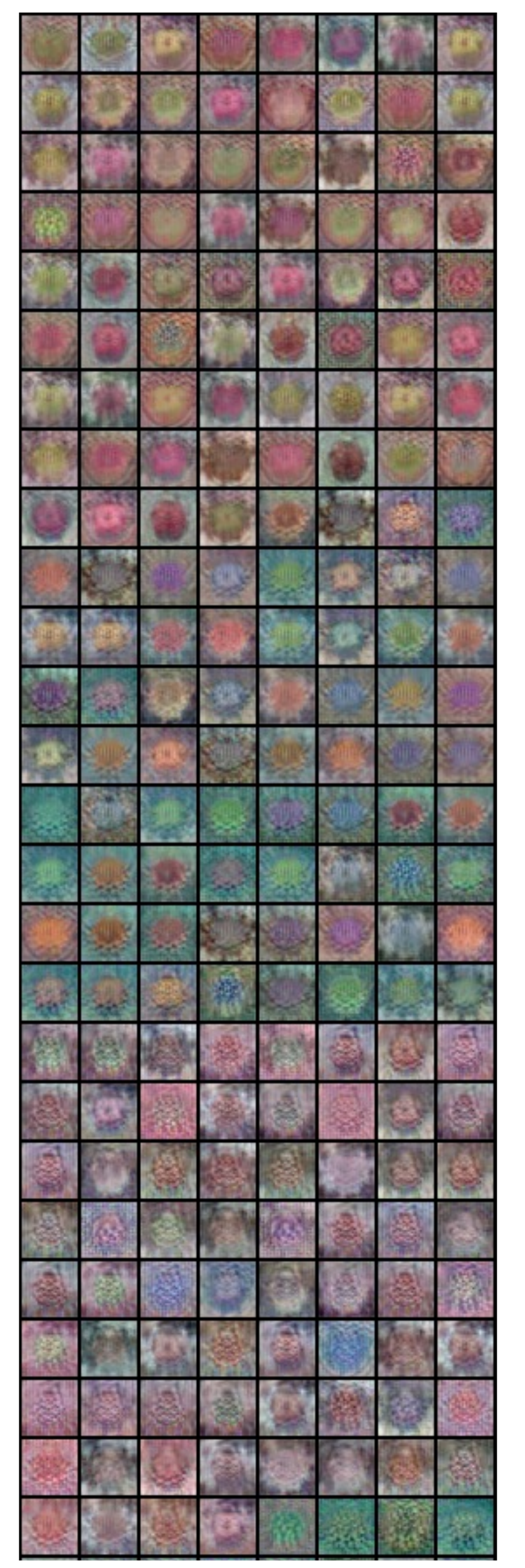}
   \caption{Recalled synthetic images for classes apple, aquarium fish, and baby under 2 I/C with 8 bases.}
   \label{fig:2ic}
\end{figure}

\begin{figure}[h]
  \centering
  \includegraphics[width=0.4\linewidth]{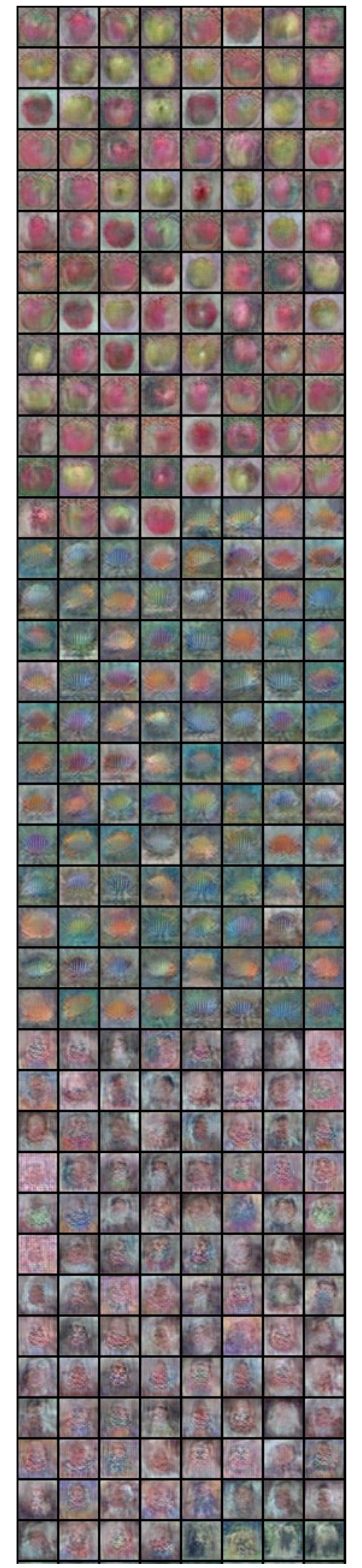}
   \caption{Recalled synthetic images for classes apple, aquarium fish, and baby under 10 I/C with 43 bases.}
   \label{fig:10ic}
\end{figure}

\begin{figure}[h]
  \centering
  \includegraphics[width=0.4\linewidth]{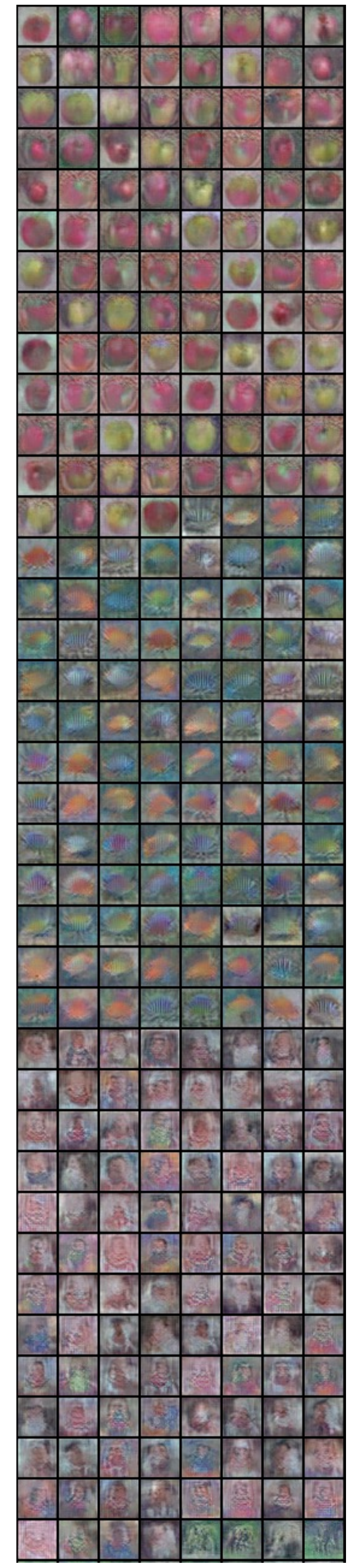}
   \caption{Recalled synthetic images for classes apple, aquarium fish, and baby under 50 I/C with 215 bases.}
   \label{fig:50ic}
\end{figure}

We compare the visualizations of synthetic images under various image per class (I/C) budgets. Similar to previous section~\ref{sec:vis-same-amount}, we use bases with the same shape, and compare the results under 2, 10 and 50 I/Cs. The corresponding number of bases are 8, 43 and 215. The visualization results are summarized in figure~\ref{fig:2ic}, figure~\ref{fig:10ic} and figure~\ref{fig:50ic}, 

\end{document}